\documentclass[onefignum,onetabnum]{siamonline190516}

\usepackage[utf8]{inputenc} 
\usepackage[T1]{fontenc}    
\usepackage{hyperref}       
\usepackage{url}            
\usepackage{booktabs}       
\usepackage{amsfonts}       
\usepackage{nicefrac}       
\usepackage{microtype}      
\usepackage{multirow}
\usepackage{booktabs}
\usepackage{subcaption}
\usepackage{wrapfig}

\usepackage{amsmath}
\usepackage{amssymb}
\usepackage{enumitem}
\usepackage{color}
\usepackage{graphicx}
\usepackage{float}
\usepackage{bm}
\usepackage{bbm}
\usepackage{circledsteps}
\usepackage[capitalize]{cleveref}
\usepackage[normalem]{ulem}

\newcommand{\iid}{i.i.d.}

\newcommand{\circled}[1]{\textcircled{\raisebox{-0.9pt}{#1}}}

\newcommand{\nrm}[1]{\textcolor[RGB]{0,0,0}{#1}}

\newcommand{\bst}[1]{\textcolor[RGB]{27,158,119}{\textbf{#1}}}

\newcommand{\stkout}[1]{\ifmmode\text{\sout{\ensuremath{#1}}}\else\sout{#1}\fi}

\newcommand{\phantomsubfigure}[1]{\begin{subfigure}[b]{0.0\textwidth}\phantomcaption\label{#1}\end{subfigure}}

\newcommand{\xhdr}[1]{\noindent{\textbf{#1.}}\hspace{0.5mm}}

\newcommand{\comment}[1]{}

\usepackage{hyperref}
\definecolor{mylinkcolor}{RGB}{0,0,170}
\hypersetup{colorlinks,allcolors=mylinkcolor,citecolor=mylinkcolor}

\usepackage{amsfonts}
\usepackage{amsmath}
\usepackage{mathrsfs} 
\usepackage{amssymb}
\usepackage{graphicx}
\usepackage{booktabs}
\usepackage{enumitem}
\usepackage{algorithm}
\usepackage{algorithmicx}
\usepackage{algpseudocode}
\usepackage{mathtools}

\newcommand{\algoid}{\mathcal{A}}

\newcommand{\mat}[1]{\bm{#1}}
\newcommand{\vct}[1]{\bm{#1}}

\newcommand{\rmat}[1]{\boldsymbol{\mathrm{#1}}}%
\newcommand{\rvct}[1]{\boldsymbol{\mathrm{#1}}}

\DeclareMathOperator*{\minimize}{minimize}

\DeclareMathOperator*{\subjectto}{subject\;to}

\DeclareMathOperator{\trace}{tr}
\DeclareMathOperator{\diagonal}{diag}
\DeclareMathOperator{\vectorize}{vec}

\DeclareMathOperator{\prob}{\rho}

\newcommand{\eye}{\mat{I}}
\newcommand{\mA}{\ensuremath{\mat{A}}}

\newcommand{\mD}{\ensuremath{\mat{D}}}

\newcommand{\mF}{\ensuremath{\mat{F}}}

\newcommand{\mH}{\ensuremath{\mat{H}}}
\newcommand{\mI}{\ensuremath{\mat{I}}}
\newcommand{\mJ}{\ensuremath{\mat{J}}}

\newcommand{\mL}{\ensuremath{\mat{L}}}

\newcommand{\mN}{\ensuremath{\mat{N}}}

\newcommand{\mS}{\ensuremath{\mat{S}}}

\newcommand{\mV}{\ensuremath{\mat{V}}}
\newcommand{\mW}{\ensuremath{\mat{W}}}
\newcommand{\mX}{\ensuremath{\mat{X}}}

\newcommand{\mZ}{\ensuremath{\mat{Z}}}
\newcommand{\mzero}{\ensuremath{\mat{0}}}

\newcommand{\mSigma}{\ensuremath{\mat{\Sigma}}}
\newcommand{\mGamma}{\ensuremath{\mat{\Gamma}}}
\newcommand{\mLambda}{\ensuremath{\mat{\Lambda}}}

\newcommand{\mTheta}{\ensuremath{\mat{\Theta}}}

\newcommand{\rmA}{\ensuremath{\rmat{A}}}

\newcommand{\rmX}{\ensuremath{\rmat{X}}}

\newcommand{\zeros}{\ensuremath{\vct{0}}}
\newcommand{\ones}{\ensuremath{\vct{1}}}
\newcommand{\va}{\ensuremath{\vct{a}}}

\newcommand{\vc}{\ensuremath{\vct{c}}}

\newcommand{\vf}{\ensuremath{\vct{f}}}
\newcommand{\vg}{\ensuremath{\vct{g}}}
\newcommand{\vh}{\ensuremath{\vct{h}}}

\newcommand{\vr}{\ensuremath{\vct{r}}}

\newcommand{\vw}{\ensuremath{\vct{w}}}
\newcommand{\vx}{\ensuremath{\vct{x}}}
\newcommand{\vy}{\ensuremath{\vct{y}}}
\newcommand{\vz}{\ensuremath{\vct{z}}}

\newcommand{\rva}{\ensuremath{\rvct{a}}}

\newcommand{\rvx}{\ensuremath{\rvct{x}}}
\newcommand{\rvy}{\ensuremath{\rvct{y}}}
\newcommand{\rvz}{\ensuremath{\rvct{z}}}

\newcommand{\ry}{\ensuremath{\mathrm{y}}}

\newcommand{\fz}{\footnotesize}
\newcommand{\cz}{\scriptsize}
\newcommand{\sz}{\small}

\crefname{section}{Section}{Sections}
\crefname{subsection}{Section}{Sections}

\title{A Unifying Generative Model for Graph Learning Algorithms: \\
Label Propagation, Graph Convolutions, and Combinations\thanks{The source code, data, and experiments are available at \url{https://github.com/000Justin000/GaussianMRF/}
\funding{ARO Award W911NF19-1-0057, ARO MURI, NSF Award DMS-1830274, and JP Morgan Chase \& Co.}}}

\author{Junteng Jia\thanks{Department of Computer Science, Cornell University (\email{jj585@cornell.edu}, \email{arb@cs.cornell.edu}).} 
\and Austin R. Benson\footnotemark[2]}

\begin{document}

\maketitle



\begin{abstract}
Semi-supervised learning on graphs is a widely applicable problem in network science and machine learning.
Two standard algorithms --- label propagation and graph neural networks --- both operate by repeatedly passing information along edges, the former by passing labels and the latter by passing node features, modulated by neural networks.
These two types of algorithms have largely developed separately,
and there is little understanding about the structure of network data that would make one of these approaches work particularly well compared to the other
or when the approaches can be meaningfully combined.
Here, we develop a Markov random field model for the data generation process of node attributes,
based on correlations of attributes on and between vertices,
that motivates and unifies these algorithmic approaches.
We show that label propagation, a linearized graph convolutional network, and their combination can all be derived as conditional expectations under our model.
In addition, the data model highlights problems with existing graph neural networks (and provides solutions),
serves as a rigorous statistical framework for understanding issues such as over-smoothing,
creates a testbed for evaluating inductive learning performance, 
and provides a way to sample graphs attributes that resemble empirical data.
We also find that a new algorithm derived from our data generation model, which we call a Linear Graph Convolution,
performs extremely well in practice on empirical data, and provide theoretical justification for why this is the case.
\end{abstract}

\section{Label propagation, graph neural networks, and estimating attributes on nodes}
Graphs, which consist of a set of nodes along with a set of edges that each connect two nodes,
are natural abstractions of relational data and systems with interacting
components. A common machine learning problem, often called \emph{graph-based
semi-supervised learning}, is to predict labels or outcomes on a subset of
nodes, given observations on the others~\cite{Zhu_2003,gallagher2008using,xu2010empirical,bilgic2010active,de2013influence,gleich2015using,kyng2015algorithms,Kipf_2016,hamilton2017inductive,peel2017graph}.
For example, a social networking company could use a user's age to better serve content; however, the age of some users might not be provided but could be estimated.
Similarly, a city planner could predict traffic on
streets in a road network given sensor readings on a subset of streets, or a
political consultant might want to forecast election outcomes in
geographically-connected regions (e.g., U.S.\ counties)
but only have polling data in certain locations.

One approach to these problems is based on the fact that two nodes connected by an
edge are often similar, a principle known as homophily in the
case of social networks~\cite{mcpherson2001birds} or assortativity more
generally~\cite{newman2003mixing}.
In the above examples,
two friends on an online social network are more likely to be of a similar age~\cite{ugander2011anatomy},
many cars on one street increases traffic on connected streets, and 
regional voting patterns are often positively correlated spatially~\cite{fernandez2014voter}.
If we only have outcomes and no other information, \emph{label propagation (LP)} algorithms
constitute a standard class of methods for making predictions at unlabeled nodes~\cite{Zhu_2003,gleich2015using,Zhou_2004}.
These algorithms find a label assignment for all nodes that 
(i) agrees with the known labels and
(ii) varies smoothly over the graph.
These algorithms can be formalized as an optimization problem of the form
\begin{equation}
\underset{\vf \in \mathbb{R}^n}{\minimize}\;\; \mu \cdot \| \vf - \vy^{(0)} \|^{2} + R(\vf),
\label{eq:lp_basic}
\end{equation}
where $\vf$ is the estimated outcome on each node,
$y_{u}^{(0)} = y_{u}$ is the original outcome at $u$ if $u$ is labeled and 0 otherwise,
$\mu$ is a constant, and
$R$ is a graph-based smoothness penalty such as $R(\vf) = \vf^{\intercal} \mL \vf$ for the combinatorial graph Laplacian $\mL$~\cite[Section 6]{zhu2005semi}.
The names label propagation, label spreading, or diffusion for graph-based semi-supervised learning
stem from algorithms that optimize objectives like those in \cref{eq:lp_basic},
which can be viewed as procedures that propagate, spread, or diffuse outcomes on labeled nodes to the unlabeled nodes.

Of course, we often have additional information or \emph{features} on nodes that are useful for estimating outcomes.
For instance, the content consumed by a user in a social network could correlate with age,
the weather affects traffic in a road network,
and unemployment levels might matter for elections.
\emph{Graph neural networks (GNNs)} are a popular class of algorithms that use such information~\cite{scarselli2008graph,Hamilton-2017-representation}.
In general, these algorithms solve the following optimization problem
\begin{align}
\textstyle \underset{\theta}{\text{minimize}}\;\; \sum_{u \in L} \left[{y}_{u} - g\left({\vx}_{u}, \{{\vx}_{v}: v \in N_{K}(u)\}, \theta\right)\right]^2,
\label{eq:gnn_basic}
\end{align}
where $N_{K}(u)$ is the $K$-hop neighborhood of vertex $u$,
$L$ is the set of labeled nodes,
${\vx}_{u}$ represents the set of features of node $u$,
$g$ is a function that usually involves neural networks, and
$y_u$ is the estimated outcome at node $u$.
The squared loss in \cref{eq:gnn_basic} is for regression (on which we focus),
although these methods are often used with a cross-entropy loss for classification problems.

The structure of \cref{eq:lp_basic,eq:gnn_basic} are markedly different.
At first glance, one might wonder why label propagation methods do not use node features.
The reason is rooted in the original motivation for many of those algorithms: semi-supervised learning for point cloud data.
In this setting, one typically assumes that 
(i) only a limited number of data points are labeled, 
(ii) the labels vary smoothly over a low-dimensional manifold, and 
(iii) the data points on where we want to make predictions are known a priori~\cite{zhu2005semi}.
The edges in the graph are then constructed from the features themselves.
More specifically, an edge encodes the similarity of two points in feature space, and the smoothness regularizer captures (ii) and (iii).
Still, label propagation methods are tremendously valuable in settings where the edges of a given graph
serve as a proxy for similarity~\cite{gleich2015using,altenburger2018monophily,eswaran2020higher,juan2020ultra,huang2020combining,tudisco2020nonlinear}.
Furthermore, early semi-supervised learning algorithms used ``external classifiers'' with label propagation~\cite{Zhu_2003}
and collective classification combines network structure and node features~\cite{sen2008collective} (see the related work below).

One might also wonder why graph neural networks do not explicit use the fact that connected nodes tend to share the same outcome or label.
As an extreme case, suppose the outcomes varies smoothly along edges in a graph where all nodes have the same features.
Then the GNN model of \cref{eq:gnn_basic} is meaningless, but the label propagation model in \cref{eq:lp_basic} is still useful.
A more realistic setting is one in which the features are mildly but not overwhelmingly predictive of the outcomes,
in which case some combination of \cref{eq:lp_basic,eq:gnn_basic} would seem reasonable. 
The machine learning community has recently developed several heuristics
that (at least implicitly) address this shortcoming:
(i) augment the features of a node with the output of a label propagation
  algorithm before using a GNN~\cite{shi2020masked};
(ii) use label propagation ideas to enforce that some intermediate
  quantity used in computing the function $g$ in \cref{eq:gnn_basic} varies
  smoothly over the graph~\cite{klicpera2018predict,bojchevski2020scaling};
(iii) use label propagation as a pre-processing step
  so that a GNN places more emphasis on certain
  connections~\cite{wang2020unifying};
(iv) use label propagation as a post-processing step to enforce smoothness of
  the estimates over the graph~\cite{huang2020combining,jia2020residual};
(v) use several functions $g$ as in \cref{eq:gnn_basic}, associating each with
  different parts of the graph where different types of outcomes may be more
  prevalent~\cite{you2019position}; or
(vi) add a random field ``layer'' on top of the GNN model to learn possible
  correlated structure in the labels~\cite{gao2019conditional,qu2019gmnn}.

These approaches are unsatisfying because they are largely ad hoc, and there is little understanding of why or when
one particular approach should work. In addition, the GNN model in
\cref{eq:gnn_basic} is opaque and not intrinsically motivated, so more
complicated methods designed on top of GNNs are unlikely to yield meaningful
insights. In general, there is limited theory for these approaches.
An underlying issue is that there has been no generative model (i.e., a stochastic data generating process) for graph data that motivates these different algorithms (like how a multivariate Gaussian model motivates linear regression).
In this paper, we address this issue with a generative model for node features and labels.
This model reveals the connections between label propagation, graph neural networks, and some of the heuristics outlined above.

\subsection{The present work: A Markov random field model for attributed graphs}

\begin{figure}[t]
\centering
\includegraphics[width=0.95\linewidth]{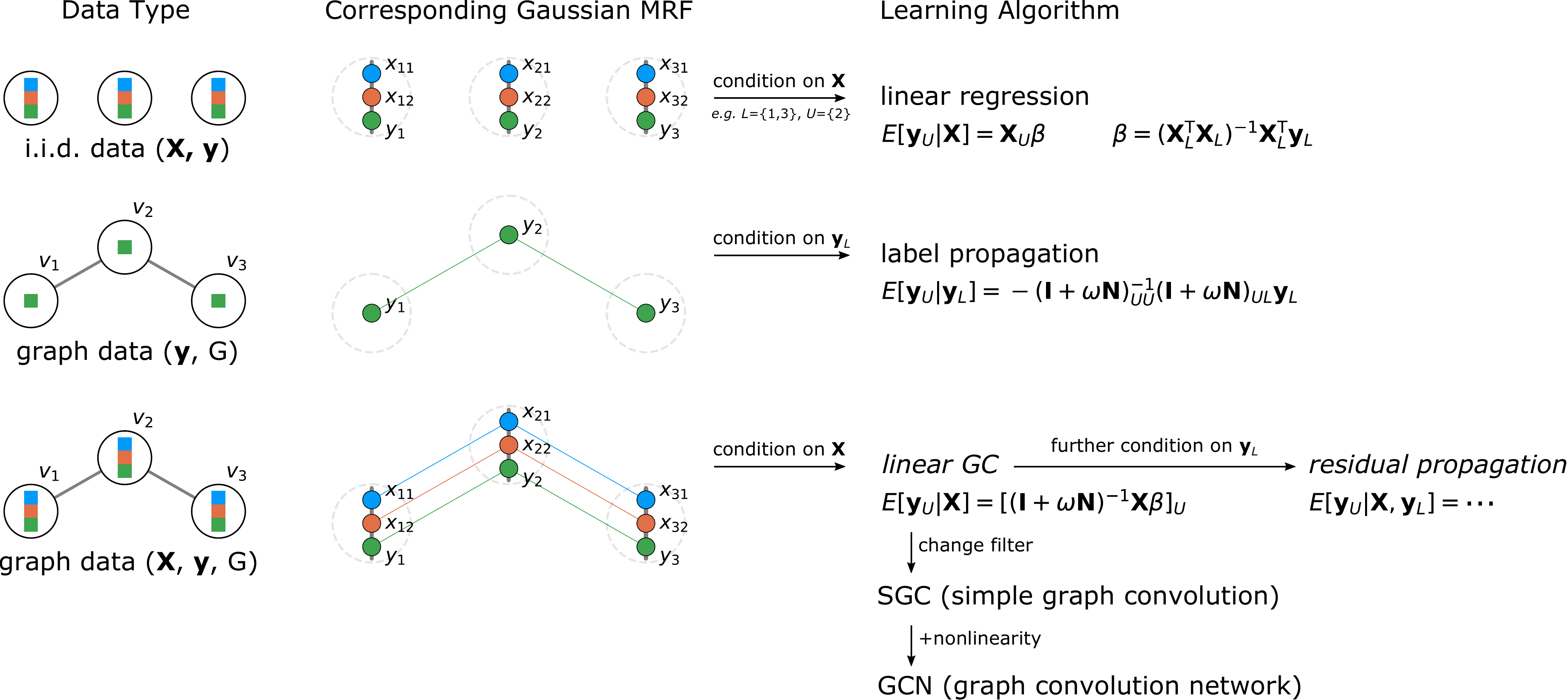}
\caption{A unified view of \iid~data (top), labeled graph data (middle), and attributed graph data (bottom) under our Gaussian MRF framework.
Each vertical bar in the Gaussian MRF diagrams connects interacting variables.
For each case, we derive a learning algorithm by the conditional expectation of the unknown labels, which gives linear regression for \iid~data, label propagation for labeled graph data, and linear graph convolution for attributed graph data.
The expectation of unknown labels conditioned on both features and observed labels leads to hybrid algorithms that combines principles from label propagation and graph neural networks.
}
\label{fig:intro}
\end{figure}

In this paper, we propose a new generative model for graph data that unifies the understanding for various graph learning algorithms.
For this, we assume that the graph topology (i.e., the nodes and edges) is given but
the \emph{attributes} --- features and outcomes --- on the nodes are random.
The main idea is to model the correlations of attributes \emph{on} and \emph{between} nodes through a random data generation process, 
after which we can understand the appropriate mechanisms to use either type of correlation (or both) to infer unknown outcomes.

More specifically, given an attributed graph, we map each attribute on each node to a random variable,
and we use a Markov random field (MRF) to describe the joint distribution of the random variables.
The MRF has two parts:
(i) potentials modeling the correlation between different attributes on the same node and 
(ii) potentials modeling the correlation between attributes among connected nodes.
By treating the outcome as one attribute and the features as the remaining attributes, the 
first type of potentials can make the features useful for estimating the outcomes.
The second type of potential has two roles.
First, they make it easy to express a smoothness in outcomes over the graph as would be assumed by label propagation methods following \cref{eq:lp_basic};
indeed, a special case case of our model when there are no features (just a single outcome attribute per node) coincides with
the Markov random field used to derive a label propagation algorithm~\cite{Zhu_2003}.
Second, the correlation in features between connected nodes makes combining features from adjacent nodes useful for estimation,
which fits the general GNN model in \cref{eq:gnn_basic} and supports the premise upon which GNNs are motivated~\cite{Hamilton-2017-representation}.

This probabilistic data model unifies the ideas behind label propagation and graph neural networks for graph data, as well as linear regression for \iid~data (\cref{fig:intro}).
In particular, if the node attributes are real-valued, then their joint distribution is a Gaussian MRF, and inference algorithms can be derived by computing the conditional expectation of the unobserved labels.
Since the conditional expectation minimizes the mean squared error of predicted labels, the derived algorithms are optimal for the quadratic objective functions commonly used for node regression (e.g., \cref{eq:lp_basic,eq:gnn_basic}).
In this sense, our analysis serves as a type of algorithmic anti-differentiation~\cite{gleich2014algorithmic} for several graph learning algorithms by showing a reasonable objective that they are not obviously optimizing.
More specifically, straightforward computations reveal the following relationships with our model:
\begin{enumerate}[leftmargin=0.2in]
\item the expected value of outcomes on unlabeled nodes, conditioned on the outcomes at labeled nodes, corresponds to a variant of label propagation;
\item the expected value of outcomes on unlabeled nodes, conditioned on the features at all nodes, corresponds to a linear approximation of the graph convolutional network (GCN)~\cite{Kipf_2016}; we call this approximation algorithm a \emph{Linear Graph Convolution (LGC)}
\item the expected value of outcomes on unlabeled nodes, conditioned on the features at all nodes and outcomes at labeled nodes
leads to the same linear GCN approximation followed by the recently proposed residual propagation heuristic~\cite{jia2020residual}.
\end{enumerate}
Finally, if there are no edges in the graph and thus no type (ii) potentials in the Gaussian MRF, then the attributes on every vertices is an \iid~sample and LGC reduces to linear regression.

Our model provides highlights when certain algorithms are suitable or helpful.
Intuitively, label propagation performs better than graph neural networks when outcome correlation (homophily)
is more influential than the correlation between features and outcomes.
Similarly, graph neural networks are more useful when correlations between features and outcomes
are stronger than homophily.
And when neither features nor homophily overwhelms the other, there is opportunity to combine ideas from label propagation and graph neural networks.

Our model also provides theoretical grounding for a number of recent empirical methodologies in graph-based machine learning.
There are several recently proposed ``simplified GNNs'' that avoid much of the neural network component~\cite{klicpera2018predict,Wu_2019,rossi2020sign,Li_2019}. 
Although these methods are algorithmically similar to our proposed LGC, they are largely motivated as computationally economic alternatives to GNNs and not as conditional expectation under some generative model.
Our data model provides some justification for why these approaches can give high prediction accuracy, but it also provides theory for why the LGC that we derive addresses fundamental shortcomings of some existing approaches.
We also find that the LGC algorithm is more accurate compared to standard GNNs on a number of empirical datasets.

The LGC algorithm arising from our analysis smooths features of nodes over the graph,
using the output of label propagation \emph{on the features} as input to a linear model.
The idea of feature smoothing arises in many contexts for understanding GNNs~\cite{bojchevski2020scaling,chien2020adaptive,Liu_2020,ma2020unified},
although some studies claim that ``over-smoothing'' is problematic~\cite{Li_2018,oono2020graph,zhao2020pairnorm}.
Our model provides a statistical view of feature smoothing
and a principled way to find an optimal smoothing level, which we can interpret
as a balance between the effects of homophily and noise.

The data model also provides a framework for evaluating \emph{inductive learning} performance, 
where a model trained on one graph is used to make predictions on another graph where the nodes have the same types of attributes.
In this setting, we can simply sample two graphs from our model, fitting parameters from one and measuring accuracy on the other.
Through this, we show how homophily can make inductive problems challenging and see how GNNs generalize poorly due
to ``memorization'' mechanisms that achieve near-zero training error.

Finally, we can fit the generative model parameters to empirical data.
From this, we can sample realistic graphs with the same attributes as test cases.

\subsection{Additional related work}
Collective classification or collective inference refers to a general set of graph-based machine learning methods that model
(i) correlations between outcome at a node and features at a node,
(ii) correlations between outcome at a node and features of neighbors of that node, and
(iii) correlation between the outcome at a node and the outcomes of neighbors of that node~\cite{jensen2004collective,sen2008collective,london2014collective}.
One class of methods called ``iterative classification'' builds consensus into a local classifier by alternating between predicting outcomes for each node with the classifier and then updating the model with the predicted outcomes on neighboring vertices~\cite{neville2000iterative,lu2003link,macskassy2007classification,mcdowell2007cautious,zheleva2009join}.
Other methods are based on a global formulation conceptually similar to this work,
where they model categorical attributes with a discrete MRF and predict unknown labels by selecting the ones with the maximum marginal probabilities.
However, since the marginally probabilities in a discrete MRF have no analytically expression, such methods require fitting the MRF potentials and then estimating the marginal probabilities with loopy belief propagation or mean field approximations~\cite{sen2008collective}; both steps are computationally expensive and require careful initialization.
In contrast, our model motivates efficient and stable learning algorithms that can be directly trained on the observed labels.

That outcomes should be correlated on a network is a longstanding idea in social network analysis~\cite{doreian1989network,leenders2002modeling,bramoulle2009identification,lin2010identifying},
where tools from spatial statistics (namely, spatial autocorrelation) are adapted to the network setting.
Recent analysis of penalty-based statistical models that encourage smoothness in parameter estimates over the graph 
provides statistical rigor to similar models~\cite{li2019prediction}.
At a high level, these models aim to estimate the effects of covariates, which
is different from our transductive setting where outcomes are missing on some nodes in the graph.



\section{A model for attributed graph data and deriving graph learning algorithms} \label{sec:methods}
Let $G = (V,E)$ be an undirected graph, where $V$ is the set of $n$ vertices (nodes) labeled $1, \ldots, n$, and $E$ is the set of $m$ edges.
We denote the (possible weighted) symmetric adjacency matrix of the graph by $\mW \in \mathbb{R}^{n \times n}$, 
and the diagonal degree matrix by $\mD \in \mathbb{R}^{n \times n}$, i.e., 
$\mD = \diagonal(\mW\ones)$, where $\ones$ is the vector of all ones.
We use $\mN = \eye - \mD^{-1/2}\mW\mD^{-1/2} = \eye - \mS$ for the normalized graph Laplacian,
where $\mS$ is the symmetrically normalized adjacency matrix.
We assume that each node $u \in V$ is associated with a vector of $p$ features $\vx_u \in \mathbb{R}^p$ and a scalar $y_u \in \mathbb{R}$ that we call the label or outcome.
We use $\mX \in \mathbb{R}^{n \times p}$ for the feature matrix and $\vy \in \mathbb{R}^{n}$ for the outcome vector.
For convenience, we stack the feature matrix and outcome vector into an attribute matrix $\mA = [\mX \; \vy] \in \mathbb{R}^{n \times (p+1)}$.
The $u$th row of the attribute matrix (denoted $\va_{u}$) represents all attributes on vertex $u$, and the $i$th column of the attribute matrix, denoted $\mA_{i}$, represents the $i$th attribute on all vertices.
For simplicity, we further assume each attribute is preprocessed to center around zero, i.e.,
$\text{mean}(\mA_{i}) = 0$ for $i = 1, \ldots, p+1$.

In this section, we consider algorithms for \emph{semi-supervised learning on graphs}, where
outcomes are available on a set of labeled nodes $L \subset V$, and we want to predict the outcome
at the remaining $U \equiv V \backslash L$ nodes.
More formally, the input to an algorithm for this problem is (i) the graph adjacency matrix $\mW$, (ii) the features $\mX$ on all nodes, and 
(iii) the labels $\vy_L$ on a subset of vertices $L \subset V$.
The output is a vector $\vy_U$ of labels on the unlabeled nodes $U \equiv V \backslash L$.
As discussed above, there are myriad algorithms for this problem.
Our goal is to develop a generative model for the node attributes that motivates and unifies several popular algorithms.

We propose a Gaussian Markov Random Field (MRF) as a generative model for attributed graph data
and show how various conditional expectations under this model lead to 
graph-based learning algorithms such as label propagation, graph convolutional networks, and residual propagation.
Since the conditional expectation minimizes the mean-squared prediction error under the data generation model,
these algorithms are ``optimal'' in some sense.
Our generative model maps each feature or label to a random variable.
We use italic symbols $\{\vx, \vy, \va, \mX, \mA\}$ for observed values of those quantities
and non-italic symbols $\{\rvx, \rvy, \rva, \rmX, \rmA\}$ for the corresponding random variables.
We assume that the graph, i.e., $\mW$, is given and not random.

Our derivations repeatedly use marginalization and conditioning of multivariate Gaussian distributions.
\Cref{subsec:gaussian_review} has the necessary background on these computations.

\subsection{The model for attributed graph data}\label{subsec:gmrf}
We assume all vertex attribute values $\mA$ are jointly sampled from a distribution over random variables $\rmA$, where each attribute on each node is a random variable.
Our model for the joint distribution of the attributes is a Gaussian Markov random field (MRF) with probability density function:
\begin{equation}
    \prob(\rmA=\mA | \mH,\vh) = \frac{e^{-\phi(\mA | \mH,\vh)}}{\int d\mA'\; e^{-\phi(\mA' | \mH,\vh)}}.
    \label{eq:gmrf}
\end{equation}
Here, the model parameter $\mH \in \mathbb{R}^{(p+1)\times(p+1)}$ is a symmetric positive definite matrix and the model parameter $\vh \in \mathbb{R}^{(p+1)}$ is entrywise positive, and the log-potential $\phi$ is defined as
\begin{align}
    \phi(\mA | \mH,\vh)
    &= \textstyle \frac{1}{2} \sum_{u=1}^{n} \va_{u}^{\intercal}\mH\va_{u} + \frac{1}{2} \sum_{i=1}^{p+1} h_{i} 
    \mA_{i}^{\intercal} \mN \mA_{i} \label{eq:gmrf_phi} \\
    &= \textstyle \frac{1}{2}\vectorize(\mA)^{\intercal}\left(\mH \otimes \mI_{n} + \diagonal(\vh) \otimes \mN\right) \vectorize(\mA).
     \label{eq:gmrf_log_potential}
\end{align}
For convenience, we write the potential as the following quadratic form:
\begin{equation}
\textstyle   \phi(\mA | \mH,\vh) = \frac{1}{2}\vectorize(\mA)^{\intercal}\, \mGamma \,\vectorize(\mA),\quad \mGamma = \mH \otimes \mI_{n} + \diagonal(\vh) \otimes \mN. \label{q:gmrf_potential_quadratic_form}
\end{equation}

At a high level, the first term in \cref{eq:gmrf_phi} encodes correlations among different attributes on each vertex.
The second term increases the probability of having attributes that are smooth over the graph, since $\vh > \mathbf{0}$
and $\mA_{i}^{\intercal} \mN \mA_{i} = \sum_{(u, v) \in E} (\nicefrac{A_{ui}}{\sqrt{d_u}} - \nicefrac{A_{vi}}{\sqrt{d_v}})^2$,
where $d_u$ denotes the degree of node $u$.
This smoothness assumption is natural for homophily.
One could choose other notions of smoothness, such as the quadratic form on the combinatorial Laplacian
instead of the normalized Laplacian; however, the normalized version will eventually lead to algorithms that are
closer to those existing in the literature.

Since  $\vh > \mathbf{0}$ and $\mH$ and $\mN$ are symmetric positive definite,
$\mGamma$ is also symmetric positive definite, and
\cref{eq:gmrf} defines the multivariate Gaussian distribution
\begin{equation}
    \prob(\mA) = (2\pi)^{-n(p+1)/2} \det(\mGamma)^{1/2} e^{-\frac{1}{2} \vectorize(\mA)^{\intercal} \mGamma \vectorize(\mA)},
\end{equation}
where $\mGamma$ is the precision matrix.
In other words, the attributes are jointly sampled via
\begin{equation}
    \vectorize(\rmA) \sim \mathcal{N}(\zeros, \mGamma^{-1}), \qquad \mGamma = \mH \otimes \mI_{n} + \diagonal(\vh) \otimes \mN.
    \label{eq:gmrf_precision}
\end{equation}
Next, we see how conditioning on observing different attributes leads to different graph learning algorithms, as illustrated in \cref{fig:intro}.

\subsection{Linear regression when there are no edges\label{subsec:lr}}
If there are no edges in the graph, then the attribute vectors over the nodes are i.i.d.\ samples from a multivariate Gaussian, resulting in a linear regression model.
This is a well-known setup for linear regression~\cite{Murphy_2012}, 
and we verify it here for our model to aid in understanding more complicated cases later.

When the edge set is empty, the log-potential function decomposes into
\begin{equation}
\textstyle \frac{1}{2}\vectorize(\mA)^{\intercal} \left(\mH \otimes \mI_{n}\right) \vectorize(\mA) = \frac{1}{2}\sum_{u=1}^{n}\va_{u}^{\intercal} \mH \va_{u},
\end{equation}
and the probability density function in \Cref{eq:gmrf} can be factorized accordingly:
\begin{equation}
    \prob(\mA)  = \frac{e^{-\frac{1}{2} \sum_{u=1}^{n} \va_{u}^{\intercal} \mH \va_{u}}}{\int d\mA'\ e^{-\frac{1}{2} \sum_{u=1}^{n} \va_{u}'^{\intercal} \mH \va_{u}'}}
    = \prod_{u=1}^{n} \frac{e^{-\frac{1}{2} \va_{u}^{\intercal} \mH \va_{u}}}{\int d\va_{u}'\ e^{-\frac{1}{2} \va_{u}'^{\intercal} \mH \va_{u}'}} = \prod_{u=1}^{n} \sqrt{\frac{\det(\mH)}{(2\pi)^{q+1}}} \cdot e^{-\frac{1}{2} \va_{u}^{\intercal} \mH \va_{u}}.
\end{equation}
In other words, $\{\va_{u}\}_{u=1}^{n}$ are \iid~samples from a multivariate Gaussian with mean $\zeros$ and precision matrix $\mH$.
Therefore, for any $u \in V$, the conditional expectation of its label $\ry_{u}$ is
\begin{equation}
    E[\ry_{u} | \rmX=\mX] = E[\ry_{u} | \rvx_{u} = \vx_{u}] = \vx_{u}^{\intercal} (-\mH_{1:p,p+1}/H_{p+1,p+1}) = \vx_{u}^{\intercal} \beta,
    \label{eq:lr}
\end{equation}
where the second equality directly follows from \cref{eq:multivariate_gaussian_conditioning_precision}.
Or, in matrix notation,
\begin{equation}
    E[\rvy|\rmX = \mX] = \mX \beta,
    \label{eq:lr_mat}
\end{equation}
which is just linear regression with coefficients $\beta = -\nicefrac{\mH_{1:p,p+1}}{H_{p+1,p+1}}$.

Rather than fitting the model parameters $\mH, \vh$, this derivation suggests a simple algorithm for label prediction --- find the optimal $\beta$ on the training data $\{(\vx_{u}, y_{u})\}_{u \in L}$ with, e.g., ordinary least squares, and make predictions $y_{u} = \vx_{u}^{\intercal} \beta$ for all $u \in U$.
There are many similar ways to arrive at linear regression and ordinary least squares,
but this pattern will be helpful for deriving the algorithms in the following sections.

\subsection{Label propagation when conditioning on observed labels\label{subsec:lp}}
Next, we consider the setting of no features ($p=0$) to demonstrate how our data model encodes label homophily and leads to label propagation (LP).
Gaussian random fields were used to develop early LP algorithms for graph learning~\cite{Zhu_2003};
our potential function is similar, so it is unsurprising that we arrive at a similar algorithm.
However, our model has a parameter that balances label homophily and noise,
which will be crucial for our graph neural network approximations.

In this setting, the positive vector $\vh$ and positive definite matrix $\mH$ reduce to positive scalars $h$ and $H$,
and our model jointly samples labels from a multivariate Gaussian:
\begin{equation}
    \rvy \sim \mathcal{N}(\zeros, \mGamma^{-1}), \qquad \mGamma = H \mI_{n} + h \mN
    \label{eq:gmrf_lp}
\end{equation}
The parameter $h$ controls homophily:
when $h$ is larger, sampled outcomes are smoother along the graph.
The parameter $H$ controls noise:
$H^{-1}$ is the variance for the outcome on each node if the graph contains no edges.
The conditional distribution of $\rvy_{U}$ given $\rvy_{L} = \vy_{L}$ is
\begin{equation}
    \rvy_{U} | \rvy_{L} = \vy_{L} \sim \mathcal{N}(\bar{\vy}_{U}, \mGamma_{UU}^{-1}),
    \label{eq:lp_conditional}
\end{equation}
where the conditional mean is
\begin{align}
    \bar{\vy}_{U} = -\mGamma_{UU}^{-1} \mGamma_{UL} \vy_{L} &= -(H \mI_{n} + h \mN)_{UU}^{-1} (H \mI_{n} + h \mN)_{UL} \vy_{L} \\
    &= -\left(\mI_{n} + \omega \mN\right)_{UU}^{-1} \left(\mI_{n} + \omega \mN\right)_{UL} \vy_{L},
    \label{eq:lp_conditional_mean}
\end{align}
with $\omega = \nicefrac{h}{H}$ as a parameter that controls the smoothing level, as we will show next.
We prove in \Cref{subsec:lp_constrained} that the conditional mean is the fixed point (i.e., $\forall u \in U, \enskip \bar{y}_{u} = y_{u}^{(\infty)}$) of the following LP algorithm
\begin{align}
    &\forall u \in U, \enskip y_{u}^{(t+1)} \leftarrow \textstyle (1 - \alpha) \cdot y_{u}^{(0)} + \alpha \cdot d_{u}^{-\frac{1}{2}} \sum_{v \in N_1(u)} d_{v}^{-\frac{1}{2}} y_{v}^{(t)}; &&\forall u \in L, \enskip y_{u}^{(t+1)} \leftarrow y_{u}^{(t)} \\
    &\forall u \in U, \enskip y_{u}^{(0)} = 0; &&\forall u \in L, \enskip y_{u}^{(0)} = y_{u},
\label{eq:lp_constrained}
\end{align}
where $\alpha = \nicefrac{\omega}{1+\omega} \in (0,1)$ grows monotonically with $\omega$,
and $N_1(u)$ is the 1-hop neighborhood of $u$ (i.e., the set of nodes connected to $u$ by an edge).
Since $\omega$ and $\alpha$ are bijective and strictly monotonic, we consider them interchangeably throughout the paper.
The update equation is equivalent to the LP formulation by Zhou et al.~\cite{Zhou_2004}, 
with the extra constraint that the predicted outcomes on observed vertices are fixed to their true values.

Alternatively, we can connect this to an optimization problem akin to \cref{eq:lp_basic}, which aligns with classical derivations of LP for semi-supervised learning~\cite{Zhu_2003,Zhou_2004}. Consider
\begin{equation}
    \minimize_{\vf \in \mathbb{R}^n} \quad \mathcal{Q}(\vf) = \mu \cdot \|\vf - \vy^{(0)}\|_2^{2} + \vf^{\intercal} \mN \vf  \qquad
         \subjectto \quad \vf_L = \vy_{L},
    \label{eq:lp_objective_constrained}
\end{equation}
where $\vy_{L}^{(0)} = \vy_{L}$ is given, $\vy_{U}^{(0)} = \zeros$ and $\mu = \nicefrac{1}{\omega}$.
Not surprisingly, \cref{eq:lp_objective_constrained} is also equivalent to the optimization formulation of Zhou et al.~\cite{Zhou_2004}, 
with the same constraint as in \cref{eq:lp_constrained} requiring the solution exactly matches the observed labels,
which was separately considered in a similar formulation by Zhu, Ghahramani, and Lafferty~\cite{Zhu_2003}.
Setting $\vf_{L} = \vy_{L}$ and $\mu = \nicefrac{1}{\omega}$, 
one can show that the conditional mean $\bar{\vy}_{U}$ is a stationary point solution of 
\cref{eq:lp_objective_constrained} and that the LP algorithm corresponds to a projected gradient descent algorithm.

In the LP algorithm of Zhou et al.~\cite{Zhou_2004}, the factor $\alpha$ just comes from 
the regularization hyperparameter in the optimization framework. 
Under our model, the optimal value of $\omega = \nicefrac{h}{H}$ (and hence $\alpha$) is determined by the homophily and noise 
levels in the data generation process --- higher homophily (larger $h$) and higher noise (small $H$) require more smoothing
(larger $\omega$), or equivalently, larger $\alpha$ in \cref{eq:lp_constrained}, placing more weight on neighboring nodes.
Still, when using LP, 
choosing $\omega$ via cross-validation is more practical than estimating $h$ and $H$,
which is the approach that we take in our numerical experiments.

For this derivation, we assumed no node features;
an alternative approach assumes that nodes have features and marginalizes over them before conditioning on observed labels.
This leads to a different LP algorithm, which we derive in \Cref{subsec:alt_lp}.
The algorithm does not have sparse computations, so we favor the above approach.

\subsection{Linear graph convolutions when conditioning on features\label{subsec:lgc}}
Next, we consider a graph sampled from our attributed graph model
and conditioning on the features of all nodes (but not the known labels).
While conditioning on both features and known labels is natural (and we develop that in the next section),
the setup here closely mirrors standard graph neural networks (GNNs) of the form in \cref{eq:gnn_basic}.
Thus, this section exposes a natural flaw in standard GNN approaches in any graph dataset with homophilous structure, 
namely that they ignore correlation in the labels.
Let $Q = \{1, \ldots, np\}$ and $P = \{np+1, \ldots, np+n\}$ denote the precision matrix indices for features and labels, respectively.
Given the features and using \cref{eq:multivariate_gaussian_conditioning_precision},
the conditional distribution of the labels is
\begin{equation}
    \rvy|\rmX = \mX \sim \mathcal{N}(\bar{\vy}, \mGamma_{PP}^{-1}).
    \label{eq:gmrf_lgc}
\end{equation}
The precision matrix is $\mGamma_{PP} = H_{p+1,p+1} \mI_{n} + h_{p+1} \mN$, where $h_{p+1}$ and $H_{p+1,p+1}$ controls the homophily level and the noise level of the outcomes.
The mean $\bar{\vy}$ of the distribution is
\begin{align}
    \bar{\vy} = E[\rvy | \rmX=\mX] &= -\mGamma_{PP}^{-1} \mGamma_{PQ} \vectorize(\mX) \\
    &= (H_{p+1,p+1} \mI_{n} + h_{p+1} \mN)^{-1} (-\mH_{1:p,p+1}^{\intercal} \otimes \mI_{n}) \vectorize(\mX) \\
    &= (\mI_{n} + \omega \mN)^{-1} \mX \beta,
    \label{eq:lgc}
\end{align}
where $\omega = \nicefrac{h_{p+1}}{H_{p+1,p+1}}$ and $\beta = -\nicefrac{\mH_{1:p,p+1}}{H_{p+1,p+1}}$.
As with linear regression, we can translate the conditional expectation into
an algorithm by fitting $\beta$ with ordinary least squares using the known labels $\vy_L$.
We call this a \emph{linear graph convolution (LGC)} based on our analysis below.

\xhdr{Comparison with linear regression}
The conditional expectation boils down to a linear function of $\beta$ in \cref{eq:lgc}.
Thus, we still reduce to linear regression, and the only difference with \cref{eq:lr_mat} is that 
the features are transformed in a pre-processing step: $\mX \to \left(\mI_{n} + \omega \mN\right)^{-1} \mX$.
When there are no edges in the graph, $\mN = \mzero$ and we reduce to \cref{eq:lr_mat}.

\xhdr{Relationship to label propagation and feature smoothing}
The transformed feature matrix $\bar{\mX} = \left(\mI_{n} + \omega \mN\right)^{-1} \mX$ resembles the LP
fixed point of \cref{eq:lp_conditional_mean}. 
This is unsurprising given that the features are also drawn from a multivariate Gaussian;
this time, we just don't have any known labels that should remain fixed.
More formally, let $\vx_{u}$ be a row of $\mX$ and $\bar{\vx}_{u}$ the corresponding row of $\bar{\mX}$;
these are the original and transformed features for node $u$.
Then, following the same steps as we did for LP,
we can write $\bar{\mX}$ as 
\begin{equation}
\bar{\mX} = \arg\min_{\mF} \| \mF - \mX \|_{F}^2 + \omega \cdot \trace(\mF^{\intercal}\mN\mF)
\label{eq:feat_transf_opt}
\end{equation}
or as the fixed-point solution of a feature propagation algorithm
\begin{align}
&\forall u \in V, \enskip \vx_{u}^{(t+1)} \leftarrow \textstyle (1 - \alpha) \cdot \vx_{u}^{(0)} + \alpha \cdot d_{u}^{-1/2} \sum_{v \in N_1(u)} d_{v}^{-1/2} \vx_{v}^{(t)} \\
&\forall u \in V, \enskip \vx_{u}^{(0)} = \vx_{u},\quad \bar{\vx}_{u} = \vx_{u}^{(\infty)},
\label{eq:feat_transf_lp}
\end{align}
where $\omega = \nicefrac{\alpha}{1-\alpha}$ for $0 < \alpha < 1$ or $\alpha = \nicefrac{\omega}{1+\omega}$ for $\omega > 0$,
and we have shifted the parameter $\mu = \nicefrac{1}{\omega}$ in the first term of the objective in \cref{eq:lp_objective_constrained} to the second term
in the objective of \cref{eq:feat_transf_opt}.

\Cref{eq:feat_transf_opt,eq:feat_transf_lp} both highlight how each column of $\bar{\mX}$ is smooth over the graph,
and the parameter $\omega$ controls the amount of smoothing, as dictated by the generative data model.
From the perspective of graph signal processing, similar
smoothing has been viewed as a 
low-pass filter for features~\cite{Tremblay_2018,Li_2019}, and
feature smoothing is a heuristic for several
graph-based learning methods~\cite{bojchevski2020scaling,chien2020adaptive,Liu_2020,ma2020unified}.
Our model puts these ideas on firm ground.

Some studies argue that feature smoothing is problematic for graph-based learning~\cite{Li_2018,zhao2020pairnorm} (the so-called
over-smoothing problem), although such analyses are largely empirical.
Our data model elucidates the problem clearly from two fronts --- attribute homophily and noise --- as captured by the parameter $\omega$.
In our model, ``oversmoothing'' is really just a result of a misspecified $\omega$.
We will see this in our numerical experiments in \Cref{subsec:smoothing}.

\xhdr{Connections to (simplified) graph convolutional networks}
To illustrate the connection of LGC to graph convolutional networks, 
we start with the fact that \cref{eq:lgc} is linear regression with feature smoothing pre-processing.
Therefore, it is natural to consider other types of feature smoothing.
For one example, the \emph{simple graph convolution} (SGC)~\cite{Wu_2019} uses the smoothing matrix 
$\tilde{\mS}^{K}$ instead of $(\mI_{n} + \omega \mN)^{-1}$, where 
$\tilde{\mS} = (\mD + \mI)^{-1/2} (\mW + \mI) (\mD + \mI)^{-1/2}$ is the normalized adjacency matrix of the graph with self-loops:
\begin{equation}
\vy^{\rm SGC} = \tilde{\mS}^{K} \mX \beta.
\label{eq:sgc}
\end{equation}
The map $\mX \to \tilde{\mS}^{K} \mX$ explicitly performs $K$ steps of weighted averaging over neighbors:
\begin{align}
    &\forall u \in V,\enskip \vx_{u}^{(t+1)} \leftarrow \textstyle (d_{u}+1)^{-1} \vx_{u}^{(t)} + (d_{u}+1)^{-1/2} \sum_{v \in N_1(u)} (d_{v}+1)^{-1/2} \vx_{v}^{(t)} \\
    &\forall u \in V,\enskip \vx_{u}^{(0)} = \vx_{u}, \quad \bar{\vx}_{u} = \vx_{u}^{(K)}.
    \label{eq:feat_transf_sgc}
\end{align}

There are a couple of mathematical oddities of SGC.%
\footnote{These oddities are artifacts of the derivation of SGC, which comes from simply
replacing the nonlinear activation functions of the Graph Convolutional Network~\cite{Kipf_2016} with linear ones.
With this formulation, Wu et al.~\cite{Wu_2019} demonstrated that nonlinearities are not essential for good empirical performance on many benchmarks.}
First, the $K$th power puts all of the emphasis on paths of length $K$,
and the self-loops are a workaround to incorporate information on shorter paths.
This contrasts with common diffusions (e.g., PageRank or Katz) that put more weight on shorter paths, often with geometric decay.
Our feature smoothing has the same geometric decay property, given by the Neumann expansion
\begin{equation}
\textstyle\left(\mI_{n} + \omega \mN\right)^{-1} = \frac{1}{1 + \omega}\left(\mI_{n} - \frac{\omega}{1 + \omega} \mS\right)^{-1} 
= (1 - \alpha)\left(I + \alpha\mS + \alpha^2\mS^2 + \ldots \right). \label{eq:neumann_series}
\end{equation}
Second, since $\| \tilde{\mS} \|_2 \leq 1$, $\tilde{\mS}^{K}$ converges to a projector onto the dominant eigenspace of  $\tilde{\mS}$ as $K \to \infty$.
In the typical case where this eigenspace is unidimensional, each column of $\tilde{\mS}^{K} \mX$ converges to the dominant eigenvector of $\tilde{\mS}$ (up to scaling), which is independent of the features!

Oddities aside, comparing \cref{eq:feat_transf_lp} with \cref{eq:feat_transf_sgc} reveals a more fundamental and important difference.
To control smoothing, our LGC uses a real-valued parameter $\alpha$, while SGC uses an integer-valued parameter $K$, (the number of layers).
This is a drawback of SGC that prevents it from achieving optimal feature smoothing, which we will see in \Cref{subsec:transductive_synthetic_results}.

The Graph Convolutional Network (GCN)~\cite{Kipf_2016}, the standard-bearer GNN, 
can be derived by adding a nonlinear transformation of the features between each application of $\tilde{\mS}$:
\begin{equation}
\vy^{\rm GCN} = \sigma(\tilde{\mS}\ldots\sigma(\tilde{\mS} \mX \mTheta^{(1)})\ldots\mTheta^{(K)}) \beta_{0},
\label{eq:gcn}
\end{equation}
where $\sigma\colon \mathbb{R}^n \to \mathbb{R}^n$ is typically the rectified linear unit (ReLU), i.e., $\sigma(\vx) = \max(\vx, \zeros)$ entrywise.
The model parameters $\mTheta^{(k)}, \beta_{0}$ are fit by minimizing $\|\vy_{L}^{\rm GCN} - \vy_{L}\|^{2}_2$.
Using a linear activation function $\sigma_{\rm id}(\vx) = \vx$, \cref{eq:gcn} collapses to \cref{eq:sgc} with $\beta = \left(\prod_{i=1}^{K} \mTheta^{(i)}\right) \beta_{0}$.
The linear feature transformation $\mX \to \tilde{\mS}^K\mX$ works empirically on some tasks,
and this approach was designed as a way to show that nonlinear activations in GCNs may be unnecessary~\cite{Wu_2019}.
While we will later compare the performances of SGC and GCN on several datasets, 
here we have given them new meanings as different approximations for the conditional expectation in our data model. 

Recall that we only conditioned on the features and not on the labels.
The known labels are just used to fit the model parameters.
Of course, if we assume that the labels are correlated on the graph, we should use that information,
which leads to the approaches in the next section.

%

%

\subsection{Residual propagation when conditioning on both features and observed labels\label{subsec:residual_propagation}}
The previous section showed that when just conditioning on observing the features $\mX$ on the nodes in our data model,
we should first smooth them with the transformation $\mX \to \left(\mI_{n} + \omega \mN\right)^{-1} \mX$.
Graph neural networks operate similarly, constructing some representation matrix $\mZ$ to use as features for a linear model
(e.g., $\mZ = \sigma(\tilde{\mS}\sigma(\tilde{\mS}\mX\mTheta^{(1)})\mTheta^{(2)})$ or $\mZ = \tilde{\mS}^2\mX$ for a 2-layer GCN or SGC).
The representation depends on labels through an optimization process as in \cref{eq:gnn_basic}.
Still, in either case, once the representations are given, predictions are independent at each node, as the objective in \cref{eq:gnn_basic} is separable.
This is strange as we often assume that outcomes are strongly correlated over the edges of a graph;
GNNs capture this implicitly at best, if at all.
Independent predictions (conditioned on representations) are a major shortcoming of GNNs,
which has led to several heuristics for incorporating labels~\cite{shi2020masked,klicpera2018predict,wang2020unifying,jia2020residual,you2019position,gao2019conditional,qu2019gmnn}.

Addressing this issue is simple within our model --- we just compute the expected label at unlabeled nodes
conditioning on \emph{both} the features on all nodes \emph{and} the known labels.
Recall that the joint distribution of the labels on all nodes given all the features is
\begin{equation}
\rvy|\rmX = \mX \sim \mathcal{N}(\bar{\vy}, \bar{\mGamma}^{-1}), \;\;\text{where}\; \bar{\vy} = (\mI_{n} + \omega \mN)^{-1} \mX \beta, \;\textrm{and}\; \bar{\mGamma} = H_{p+1,p+1} \mI_{n} + h_{p+1} \mN.
\end{equation}

Using \cref{eq:multivariate_gaussian_conditioning_precision}, the conditional distribution of $\rvy_{U}$ given 
the observed labels $\rvy_{L} = \vy_{L}$ is
\begin{equation}
    \rvy_{U} | \rmX = \mX, \rvy_{L} = \vy_{L} \sim \mathcal{N}\left(\bar{\vy}_{U} - \bar{\mGamma}_{UU}^{-1} \bar{\mGamma}_{UL} (\vy_{L} - \bar{\vy}_{L}), \bar{\mGamma}_{LL}^{-1}\right).
\end{equation}
The mean of this distribution is also the conditional expectation of $\rvy_{U}$:
\begin{align}
    E[\rvy_{U}|\rmX = \mX, \rvy_{L} = \vy_{L}] 
    &= \bar{\vy}_{U} - \bar{\mGamma}_{UU}^{-1} \bar{\mGamma}_{UL} (\vy_{L} - \bar{\vy}_{L})           \\
    &= \bar{\vy}_{U} - (\mI + \omega \mN)_{UU}^{-1} (\mI + \omega \mN)_{UL} \vr_{L} 
    = \bar{\vy}_{U} + \bar{\vr}_{U},
    \label{eq:residual_propagation}
\end{align}
where $\omega = \nicefrac{h_{p+1}}{H_{p+1,p+1}}$, 
$\vr_{L} = (\vy_{L} - \bar{\vy}_{L})$ is the regression residual on the observed labels,
and we define $\bar{\vr}_{U} = -(\mI + \omega \mN)_{UU}^{-1} (\mI + \omega \mN)_{UL} \vr_{L}$.
This is almost the conditional expectation in \cref{eq:lp_conditional_mean}, 
which was the fixed point of an LP algorithm.
The difference is that the \emph{residuals} are used for initialization, i.e.,
$\bar{\vr}_U$ is the fixed point of the following propagation algorithm
\begin{align}
    &\forall u \in U, \enskip r_{u}^{(t+1)} \leftarrow \textstyle (1 - \alpha) \cdot r_{u}^{(0)} + \alpha \cdot d_{u}^{-\frac{1}{2}} \sum_{v \in N_1(u)} d_{v}^{-\frac{1}{2}} r_{v}^{(t)}; &&\forall u \in L, \enskip r_{u}^{(t+1)} \leftarrow r_{u}^{(t)} \\
    &\forall u \in U, \enskip r_{u}^{(0)} = 0; &&\forall u \in L, \enskip r_{u}^{(0)} = y_{u} - \bar{y}_{u},
\label{eq:rp_constrained}
\end{align}
where $\nicefrac{\alpha}{1 - \alpha} = \omega$ for $0 < \alpha < 1$.
Thus, $\bar{\vr}_{U}$ can be interpreted as the estimated regression residuals on the unlabeled vertices, which is used as a correction term for the
LGC predictions.

In summary, \cref{eq:residual_propagation} leads to a 3-step algorithm:
(i) Compute the LGC prediction $\bar{\vy}_U$ with \cref{eq:lgc}, fitting $\beta$ with known labels;
(ii) Compute the regression residual on labeled vertices $\vr_{L} = \vy_{L} - \bar{\vy}_{L}$, and 
use LP initialized with these residuals to estimate the residuals on unlabeled vertices;
(iii) Add the estimated residual to the base prediction to get a final prediction on the unlabeled vertices $U$ (\cref{eq:residual_propagation}).
We refer to steps (ii) and (iii) of this algorithm as residual propagation (RP),
and the entire algorithm as LGC/RP, since $\bar{\vy}_{U}$ is the prediction given by LGC.
Here, the RP name comes from our prior work, in which nearly the same correction step was developed as a heuristic layer on top of
several graph learning algorithms, where $\bar{\vy}_U$ is replaced with a more general predictor, such
as the output of a graph neural network or a multi-layer perceptron~\cite{huang2020combining,jia2020residual}.
Putting all of the algebra together, LGC/RP is a linear model in $\beta$ and uses only linear transformations of $\mX$:
\begin{equation}
\vy^{\rm LGC/RP}_U = [(\mI_{n} + \omega \mN)^{-1} \mX \beta]_U
 -(\mI + \omega \mN)_{UU}^{-1} (\mI + \omega \mN)_{UL} \left(\vy_{L} - [(\mI_{n} + \omega \mN)^{-1} \mX \beta]_L\right) \label{eq:lgcrp_linear},
\end{equation}
although the three-step procedure above is easier to interpret.

We will see that adding RP to other algorithms for estimating $\bar{\vy}_U$ improves empirical performance,
matching prior research~\cite{huang2020combining,jia2020residual}.
However, we emphasize that the LGC/RP algorithm was derived from simply computing a conditional expectation on all of the available
information (features and labels) in our attributed graph model.

\section{Transductive learning experiments\label{sec:transductive_learning}}
We have shown that label propagation (LP), linear graph convolution (LGC), and LGC with residual propagation (LGC/RP) 
are all conditional expectations of the unknown labels in our Gaussian MRF distribution,
and that linear regression (LR) arises from a conditional expectation ignoring the graph topology.
We now evaluate the performances of these algorithms on synthetic and empirical datasets, 
in order to understand relationships among smoothing, nonlinearities, and out-of-sample prediction performance.
This section focuses on the \emph{transductive} learning setting, 
where we train the models on some vertices in a graph and test on the remainder in the same graph.
\Cref{sec:inductive_learning} considers \emph{inductive} learning, where the models are trained and tested on vertices from different graphs.

\subsection{Experiments on synthetic datasets\label{subsec:transductive_synthetic_results}}

We start our experiments on synthetic graph attributes sampled from the Gaussian MRF model.
%
First, we select the synthetic graph topology $G(V,E)$ by sampling from a Watts-Strogatz model with $1000$ vertices, average degree of $6$, and rewiring probability of $1\%$.
We set $p=4$ and randomly initialize the Gaussian MRF parameters, creating $\mH$ as follows:
(1) uniformly sample five 5-dimensional Gaussian random vectors $\{\vz_{i}\}_{i=1}^{5}$;
(2) create a matrix $\mF \in \mathbb{R}^{5 \times 5}$ with $F_{ij} = \vz_{i}^{\intercal} \vz_{j}$; and
(3) set $\mH = (\mF + 0.01*\mI)^{-1}$.
If the vertex attributes are \iid ($\vh = 0$), then $\mF_{ij}$ --- given by the inner product of $\vz_{i}$ and $\vz_{j}$ --- is also the covariance between attribute $i$ and $j$.
For the parameter $\vh$ that controls the homophily strength of each attribute, we let $h_{i} = h_{0} \cdot 10^{b_{i}}$, where $\{b_{i}\}_{i=1}^{5}$ are sampled from a uniform distribution supported on $[-0.5,0.5)$.
We sample three attributed graphs, using $h_{0} = 1, 10, 100$, where larger $h_{0}$ values corresponds to higher level of homophily.

\xhdr{Experimental setup}
We compare LP, LGC, and LGC/RP with baselines linear regression (LR), SGC, and GCN.
We also apply residual propagation post-processing to SGC and GCN, and we call these algorithms SGC/RP and GCN/RP.
For each dataset, the vertices are randomly split into $30\%$ training and $70\%$ testing, and we use the coefficient of determination $R^{2}$ to measure regression accuracy.
Our LP implementation uses the constrained formulation \cref{eq:lp_constrained}.
The pre-processing smoothing step in LGC is performed by running the propagation algorithm of \cref{eq:feat_transf_lp} on the features.
Moreover, for the graph convolutional models SGC and GCN, we use the SAGE-GCN variant~\cite{hamilton2017inductive} that subsamples each vertex's neighborhood to lower the computational cost, with maximum subsampling size equal to $30$.
The neural network parameters are trained using the ADAMW optimizer with learning rate $10^{-3}$ and decay rate of $2.5 \times 10^{-4}$.
The hyperparameters $\alpha$ in LP, LGC, residual propagation,
number of layers $K$ in SGC and GCN, and number of training epochs
are tuned with 5-fold cross validation on the training set.

Each attributed graph has five attributes, and we use each one as either a feature or outcome in separate experiments.
In each experiment, we choose a single attribute as the outcome and use the remaining four as features for prediction.
Each combination of dataset, choice of label, and algorithm is repeated 10 times with different random splits.
Since the synthetic data generation does not distinguish different attributes, we further average the accuracies across different choices of labels for the same dataset and prediction algorithm.
The performances of different methods as well as the optimal hyperparameters are summarized in \cref{tab:synthetic_accuracy}.

\begin{table*}[t]
    \caption{Accuracy ($R^{2}$) on synthetic datasets. Each reported accuracy is averaged over 5 different choices of label, each repeated 10 times with different random data splits. We report the $\alpha$ values and number of layers $K$ that give the highest validation accuracy, averaged over the experiments. More homophily (larger $h_0$) leads to more smoothing (larger $\alpha$ or $K$).}
    \centering
    \resizebox{1.0\linewidth}{!}{
    \begin{tabular}{r @{\quad} c @{\quad} cccc @{\quad} ccc}
    \toprule
    $h_{0}$  & LP ($\alpha$)     & LR   & LGC ($\alpha$) & SGC ($K$)        & GCN ($K$)        & {\sz LGC/RP ($\alpha$)} & {\sz SGC/RP ($K, \alpha$)} & {\sz GCN/RP ($K, \alpha$)} \\
    \midrule
    $1$      & 0.19 {\cz (0.79)} & 0.68 & 0.70 {\cz (0.28)} & 0.37 {\cz (1.8)} & 0.34 {\cz (1.7)} & \bst{0.73} {\cz (0.29)}    & 0.40 {\cz (1.8, 0.21)}     & 0.37 {\cz (1.7, 0.21)}     \\
    $10$     & 0.43 {\cz (0.95)} & 0.48 & 0.58 {\cz (0.57)} & 0.45 {\cz (2.1)} & 0.45 {\cz (2.0)} & \bst{0.68} {\cz (0.56)}    & 0.56 {\cz (2.1, 0.46)}     & 0.54 {\cz (2.0, 0.43)}     \\
    $100$    & 0.59 {\cz (0.99)} & 0.24 & 0.42 {\cz (0.85)} & 0.38 {\cz (2.3)} & 0.45 {\cz (2.5)} & \bst{0.64} {\cz (0.85)}    & 0.63 {\cz (2.3, 0.81)}     & 0.62 {\cz (2.5, 0.79)}     \\
    \bottomrule
    \end{tabular}
    }
    \label{tab:synthetic_accuracy}
\end{table*}

\xhdr{Main accuracy results}
The algorithms in \cref{tab:synthetic_accuracy} can be divided into 3 groups based on the information they use: 
\circled{1} LP that only uses outcome homophily;
\circled{2} feature-based methods LR, LGC, SGC and GCN that use the correlation between (transformed) features and outcome; and
\circled{3} hybrid methods LGC/RP, SGC/RP and GCN/RP that use both.

Comparing the performances across these different groups of algorithms, different algorithms are favored by different datasets.
For data sampled with a small $h_{0}$ ($h_0 = 1$), the feature correlations are more important;
in this case, the feature-based methods are far superior to LP.
On the other hand, when $h_{0} = 100$, the outcome homophily is more influential, and LP out-performs the feature-based models.
For the intermediate value $h_{0} = 10$, LGC and LP have comparable accuracy,
and the hybrid method LGC/RP is able to improve over both algorithms by a large margin.
Finally, the hybrid methods are strictly better than both LP and their feature-only counterparts, even 
though SGC/RP and GCN/RP are just heuristic algorithms.

Within the feature-based methods (group \circled{2}), 
the feature smoothing step in LGC always provides improvement over LR.
This is not the case for SGC, which has serious performance degradation on attributes sampled with low homophily levels ($h_{0} = 1$), 
where LR is a better choice.
One may attribute the performance difference between LGC and SGC to the fact that LGC is derived from the model class from which the data is sampled.
However, we will see that LGC also performs better on empirical data,
and we provide some intuition for this shortly.

When comparing the performance within the hybrid methods (group \circled{3}), LGC/RP always gives the highest regression accuracy.
This is entirely expected given that LGC/RP computes the expected outcomes conditioned on all available information, 
and thus should be the optimal algorithm measured by the sum-of-the-squared errors (or $R^{2}$ coefficient).

\xhdr{Interpretation of optimal $\alpha, K$ smoothing parameters}
In addition to the regression accuracies, \cref{tab:synthetic_accuracy} reports the optimal hyperparameters of the algorithms.
As the homophily level increases with $h_{0}$, 
the optimal smoothing parameters $\alpha$ in LP and RP increases, which is consistent with our theory in \Cref{subsec:lp,subsec:residual_propagation}.
For feature smoothing in LGC, the optimal smoothing parameter $\alpha$ also increases with $h_{0}$, 
as the algorithms place more weight on the neighboring nodes to reduce noise.
Similarly, SGC and GCN perform better with more convolutional layers, meaning that they use larger neighborhoods to increase smoothness.
While the hyperparameter $\alpha$ in LGC can take any value between 0 and 1, 
the number of layers in SGC and GCN can only take integer values and is therefore less flexible.
This is part of the reason SGC underperforms LGC by a large margin, which we discuss next.

\xhdr{Comparing feature smoothing in LGC and SGC}
We showed in \Cref{subsec:lgc} that one way to understand the difference between LGC and SGC is that
they perform feature smoothing differently.
At first glance, featuring smoothing with SGC seems more straightforward:
while applying $(\mI + \omega \mN)^{-1}$ in LGC requires running the propagation in \cref{eq:feat_transf_lp} until convergence, 
SGC simply performs $K$ rounds of weighted averaging using $\tilde{\mS}^{K}$.
We use the graph signal processing framework~\cite{ortega2018graph}
to analyze both smoothing techniques as low-pass filters, where $(\mI + \omega \mN)^{-1}$ has much more expressivity,
as a way to understand their performance differences.

Let $\mN = \mV \mLambda \mV^{\intercal}$ denote the symmetric eigendecomposition of the normalized Laplacian matrix.
Each eigenvector $\mV_{i}$ can be interpreted as a signal over the graph, where $V_{ji}$ is the signal strength on node $j$.
Eigenvectors with small eigenvalues correspond to smooth signals (over the graph)
and eigenvectors with large eigenvalues correspond to more oscillatory signals.

Let $\vf \in \mathbb{R}^{n}$ be a column of $\mX$ representing a feature over the graph and write 
$\vf = \sum_{i} c_{i} \mV_{i} = \mV \vc$, where $c_{i}$ is the coefficient for the $i$th eigenvector.
Then, by the spectral mapping theorem,
\begin{equation}
    (\mI + \omega \mN)^{-1} \vf = \mV (\mI + \omega \mLambda)^{-1} \mV^{\intercal} \vf = \mV (\mI + \omega \mLambda)^{-1} \vc = \mV \tilde{\vc},
\end{equation}
where $\tilde{c}_{i} = \nicefrac{c_{i}}{(1+\omega\lambda_{i})}$ is the new coefficient of the processed signal for the $i$th eigenvector.
We can see $(\mI + \omega \mN)^{-1}$ suppresses the coefficients for eigenvectors with larger eigenvalues,
while leaving the coefficient for eigenvector with $\lambda_{i} = 0$ intact, so it is a low-pass filter on graph signals.
Since eigenvectors corresponding to larger eigenvalues are more oscillatory over the graph, suppressing them smooths the features.

\begin{figure}[t]
\centering
\includegraphics[width=0.90\linewidth]{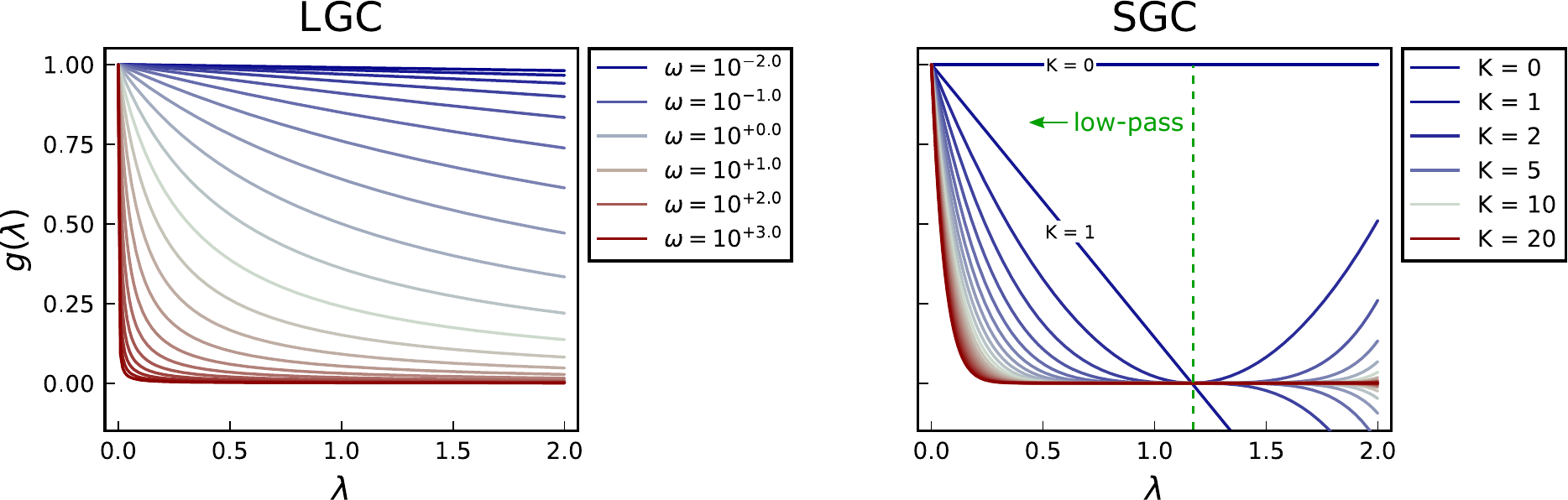}
\caption{Frequency response functions for the smoothing filters of LGC and SGC on a $6$-regular graph.
The color gradient (blue $\to$ red) corresponds to increasing ($\omega$ $10^{-2} \to 10^{3}$) and $K$ ($0 \to 20$).
There are two limitations of the SGC filter $\tilde{\mS}^{K}$:
(i) it is not low-pass for eigenvalues $\lambda_{i} \in [\nicefrac{(d+1)}{d}, 2]$ and
(ii) even on its low-pass range $[0, \nicefrac{(d+1)}{d}]$, it is not that expressive.
}
\label{fig:frequency_response}
\end{figure}

We write $\tilde{c}_{i} = c_{i} \cdot g(\lambda_{i})$, where $g(\lambda_{i}) = \nicefrac{1}{(1+\omega\lambda_{i})}$ is called the frequency response function of the LGC filter, which we will compare with the function for SGC.
For simplicity, assume that $G$ is $d$-regular, so we can
diagonalize $\tilde{\mS}^{K}$ with the basis of eigenvectors $\mV$ for $\mS$:
\begin{equation}
    \tilde{\mS}^{K} = \left((\mD+\mI)^{-\frac{1}{2}} (\mW+\mI) (\mD+\mI)^{-\frac{1}{2}}\right)^{K} = \left(\mI - \nicefrac{d}{(d+1)} \mN\right)^{K} = \mV \left(\mI - \nicefrac{d}{(d+1)} \mLambda\right)^{K} \mV^{\intercal}
\end{equation}
The corresponding frequency response function is $g(\lambda_{i}) = \left(1 - \nicefrac{d}{(d+1)} \lambda_{i}\right)^{K}$
and corresponds to a low-pass filter for eigenvalues in $[0,\nicefrac{(d+1)}{d}]$.
Prior research notes that $\tilde{\mS}^{K}$ suffers from the fact that is it not a low-pass on the entire range of eigenvalues~\cite{Wu_2019,Li_2019}, 
but we point out a more pressing issue here --- the filter is not expressive enough even on the range where it is low-pass.

The expressiveness of a low-pass filter can be measured by how well the frequency response function
can approximate an arbitrary decaying function.
Our design choices are a continuous parameter $\omega$ for LGC and an integer parameter $K$ for SGC,
which already hints at why LGC might be able to construct better filters.
We choose $d=6$ and compare the frequency response functions for LGC and SGC in \cref{fig:frequency_response}.
Tuning $\omega$ spans a large space of convex smoothly decaying functions over the eigenvalues.
In contrast, the frequency response function for $\tilde{\mS}^{0}$ and $\tilde{\mS}^{1}$ are vastly different, 
and there is no intermediate filter between them, as $K$ must takes integer values.
When smoothing features, one has to trade off the desire to reduce noise (increase smoothness) and  preserving the signal (decrease smoothness).
The LGC filter can choose $\omega$ to find the sweet spot,
whereas the SGC filter has limited flexibility; in some cases, the noise hurts accuracy without feature smoothing ($K=0$),
whereas using just $\tilde{\mS}^{1}$ may force too smooth a signal.
(\Cref{subsec:smoothing} discusses how to choose an optimal smoothing level $\omega$.)

\xhdr{Effects of non-linearities}
We analyze the effects of non-linear activation functions by comparing GCN with SGC.
The GCN performs marginally worse than the latter when $h_{0}=1$, yet considerately better when $h_{0}=100$.
This might seem strange at first glance, since linear models are optimal for our synthetic data.
However, nonlinear functions are still helpful in two ways.
First, SGC is limited as a low-pass filter, and the nonlinearities provide additional modeling power that could be helpful for approximating the right filters.
%
Second, the increased modeling power gives the GCN higher capacity to partially ``memorize'' the training examples, as evidenced by near-zero loss on the training set.
%
When the outcomes are smooth along the graph structure (larger $h_{0}$), memorizing the features in the neighborhood of each training node and the corresponding label should help out-of-sample prediction, as there are overlaps in the neighborhoods of the training and testing vertices.
This is connects to the common practice of setting $\mX = \mA$ when no features are available~\cite{Kipf_2016},
where $\mA\mTheta^{(1)}$ makes the initial-layer embedding of two nodes similar if they have similar sets of neighbors.
The memorization has downsides.
Although the GCN performs better than the SGC here in the transductive setting when $h_{0} = 100$, we will show in \Cref{sec:inductive_learning} 
that it performs worse in the inductive setting on an unseen graphs sampled under the exact same parameters, simply because the training and testing vertices no longer share any neighbors.

\subsection{Identifiability, over-smoothing, and under-smoothings\label{subsec:smoothing}}

We have made the expressibility argument for LGC (over SGC).
Next, we focus on an identifiability problem: can we recover the true $\omega$?
It turns out that standard hyperparameter tuning strategies are sufficient for synthetic data,
motivating the same strategy for empirical data.
To demonstrate this, we again use synthetic datas generated from our model, 
where the optimal smoothing level is determined by the parameter values of $\mH$ and $\vh$ used during sampling (see \Cref{sec:methods}).

We start with the LP algorithm.
Typically, the parameter $\alpha$ is relatively large, such as $\alpha = 0.99$~\cite{Zhou_2004}, which corresponds to $\omega = 99$.
This may be sufficient for the classification setting, where a ``score'' for each class is generated for each node, and all that matters for accuracy
is whether or not the correct class has the highest score.
In the regression setting, where performance is measured with squared error, choosing a good hyperparameter is crucial.
For synthetic data generation, we use the same graph topology as in \Cref{subsec:transductive_synthetic_results} and consider different parameters of our model.
We set $p = 0$ (no features), so the Gaussian MRF precision matrix reduces to $\mGamma = H \mI_{n} + h \mN$.
Since overall scaling in all the outcomes does not change the regression performance, we further simplify with $\mGamma = \mI_{n} + \omega \mN$.
We generate synthetic graphs for a range of $\omega$ values from $10^{-1}$ to $10^{2}$.
For each graph, we randomly select $30\%$ of the vertices for training and test the regression accuracy on the remaining vertices,
using estimates generated via the procedure in \cref{eq:lp_constrained} with a range of $\omega$ values.
These $\omega$ values are our guesses for the ``true $\omega$'' and the estimate is the one that gives the best out-of-sample accuracy.
\begin{figure}[t]
\phantomsubfigure{fig:demo_synthetic_a}
\phantomsubfigure{fig:demo_synthetic_b}
\phantomsubfigure{fig:demo_synthetic_c}
\centering
\includegraphics[width=0.90\linewidth]{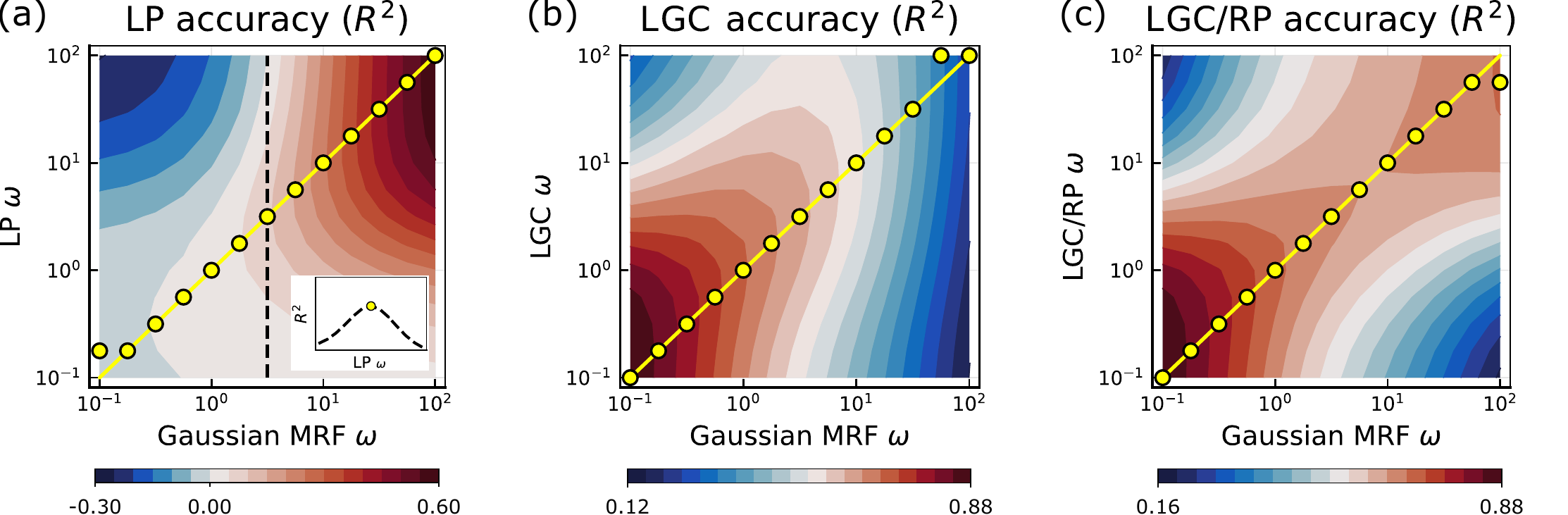}
\caption{(a) Performance of LP on synthetic attributed graphs ($p=0$).
Labels are drawn from a Gaussian MRF with precision matrix $\mathbf{I} + \omega \mN$ for various $\omega$.
For each sample, we run LP with different $\omega$ and measure the accuracy $R^{2}$.
For each Gaussian MRF $\omega$, the yellow circle marks the $\omega$ value for LP that gives the largest $R^2$.
The inset shows the performance for one $\omega$ used to generate the data, corresponding to the vertical black dashed line.
Most circles are on the diagonal yellow line, confirming that the optimal LP $\omega$ parameter is essentially the true Gaussian MRF $\omega$.
(b) Performance of LGC ($p=1$).
The attributes in each graph are drawn from a Gaussian MRF, and the optimal smoothing parameter is $\omega = h_{2}/H_{22}$.
For each sample, we run LGC with different $\omega$ parameters.
The LGC $\omega$ parameter that gives the highest $R^2$ is always close to the $\omega$ used to generate the data.
(c) The performance of LGC with residual propagation on the same synthetic attributed graphs as in (b).
Again, the LGC/RP $\omega$ parameter that gives the highest $R^{2}$ matches the optimal $\omega$ used to generate the data.
}
\label{fig:demo_synthetic}
\end{figure}

Each combination of Gaussian MRF $\omega$ (for data generation) and LP $\omega$ (for prediction) is repeated $30$ times with different random seeds, and the mean accuracy is plotted in \cref{fig:demo_synthetic_a}.
The empirical optimal LP $\omega$ under each Gaussian MRF $\omega$ is marked with a yellow dot.
We would hope that LP works the best when its $\omega$ estimate matches the Gaussian MRF $\omega$ parameter used for sampling.
This is confirmed, as most yellow dots in \cref{fig:demo_synthetic_a} are on the diagonal.
In other words, tuning the $\omega$ for LP can indeed find the optimal smoothing level.

We ran analogous experiments for LGC and LGC/RP.
In this case, the Gaussian MRF has two attributes (one feature and one label) with parameters $\vh \in \mathbb{R}^{2}$ and $\mH \in \mathbb{R}^{2 \times 2}$.
We fix the matrix elements $H_{11} = H_{22} = 1.0, H_{12} = H_{21} = -0.99$ and vary the homophily strength $h_{1} = h_{2}$ from $10^{-1}$ to $10^{2}$.
The theoretically optimal smoothing parameter for LGC and LGC/RP (given by $\omega = h_{2}/H_{22}$) also varies from $10^{-1}$ to $10^{2}$.
We again sample attributed graphs from the Gaussian MRF model, use 30\% of vertices for training and 70\% for testing,
average over 30 random samples, and compare whether the estimated $\omega$ (by selecting the $\omega$ that gives the best accuracy)
matches the $\omega$ used to generate the data.
Similar to the experiments on LP, we find the estimated $\omega$ for LGC and LGC/RP always approximately match the $\omega$ used to generate the data
(\cref{fig:demo_synthetic_b,fig:demo_synthetic_c}).

When the homophily level in the dataset is not overwhelming large (e.g., $\omega < 10^{1}$ in the generative model), 
running LGC (or LGC/RP) with a larger estimated $\omega$ causes significant performance degradation, as shown by the top-left corner of each subplot.
This corresponds to the aforementioned ``over-smoothing'' phenomenon in the practice of GNNs~\cite{Li_2018,oono2020graph,zhao2020pairnorm}. 
Our data model makes it clear when and why this could be a real problem.
The LGC model avoids over-smoothing by finding the right $\omega$ using the validation set, but
the same may not not true for other GNNs (we saw this with SGC in \cref{fig:frequency_response}).
Similarly, the bottom-right corner of the plot corresponds to an ``under-smoothing'' region, where the 
algorithms would not take full advantage of the homophily levels to denoise the sample; our attributed graph model
shows that this can happen,  but this issue has not been studied in the GNN literature.

A difference in \cref{fig:demo_synthetic_b} and \cref{fig:demo_synthetic_c} is that the absolute performance of LGC (with optimal $\omega$) decreases as the Gaussian MRF $\omega$ increases.
The reason is that the sampled graph is increasingly governed by homophily, and LGC ignores outcome correlation.
LGC/RP still performs well with large $\omega$, as it captures outcome correlation with residual propagation.

\subsection{Experiments on empirical datasets\label{subsec:transductive_realworld_results}}
\begin{wraptable}{r}{0.42\textwidth}
\vspace{-11mm}
\begin{minipage}{1.00\linewidth}
\begin{table}[H]
    \caption{Summary of datasets statistics.}
    \centering
    \resizebox{1.00 \linewidth}{!}{
    \begin{tabular}{l @{\quad} rrr}
    \toprule
    Dataset  & \# vertices &   \# edges & \# attributes \\
    \midrule
    U.S.     &       3,106 &     22,574 &           10  \\
    CDC      &       3,107 &      9,230 &            5  \\
    London   &         608 &      1,750 &            4  \\
    Twitch   &       1,912 &     31,299 &           66  \\
    \bottomrule
    \end{tabular}
    }
    \label{tab:dataset_statistics}
\end{table}
\end{minipage}
\vspace{-5mm}
\end{wraptable}
Now that we understand the performances of different algorithms on the synthetic datasets, we turn our 
attention to performance on four empirical (real-world) attributed graphs (summary statistics are in \cref{tab:dataset_statistics}).

\emph{U.S. election map.} Each vertex in this dataset represents a U.S.\ county~\cite{jia2020residual}.
    The vertex attributes are demographic information (e.g., median education level and unemployment rate) and election margins of victory
    during the 2016 presidential election.
    Each county also has three features of the average share of Facebook friends living within 50, 100, and 500 miles,
    from the Facebook social connectedness index dataset~\cite{Badger_2018,Bailey_2018}.
    The graph topology from the Facebook dataset, where each county is connected to approximately $20$ others with the strongest social ties.
    This social graph is non-planer due to long range friendships (e.g., Wayne County, OH is connected to Loving County, TX).

    \emph{CDC climate data.} Here, vertices correspond to counties in the United States and edges connect bordering counties.
    The attributes are real-valued environmental indicators including air temperature, land temperature, precipitation, sunlight, and pm2.5, averaged over 2008~\cite{NLDAS_2012}.
    
    \emph{London election map.} The vertices represent wards in London, and the edges connect adjacent wards~\cite{Godoy_2020}.
    The vertex attributes are demographic information (average age, education level and income) and election margin of victory in each ward during the 2016 mayor election.
    
    \emph{Twitch social network.} This is a friendship network of Twitch streamers in Portugal~\cite{Rozemberczki_2019}.
    The raw vertex features are binary variables such as the games played and liked, and streaming habits.
    We project these onto the first 64 principal components for 64 real-valued features.
    The graph has high degree heterogeneity, so we also include the square root of vertex degree as a feature.
    The outcome is the number of days each streamer has been streaming on Twitch.

\begin{table*}[t]
    \caption{Transductive learning accuracies on real-world datasets as measured by the coefficient of determination ($R^{2}$). Each reported accuracy is averaged over 10 runs with different random data splits.
    Optimal hyperparameters for LGC ($\alpha$) and SGC/GCN ($K$) are reported,
    and are positively correlated.
    Even though this data is not sampled from our Gaussian MRF model,
    the LGC/RP algorithm derived from that model is accurate.
    }
    \centering
    \resizebox{\linewidth}{!}{
    \begin{tabular}{r l c cccc ccc}
    \toprule
    Dataset                       & {\sz Outcome}         & LP         & LR         & LGC ($\alpha$)       & SGC ($K$)              & GCN ($K$)              & {\sz LGC/RP}  & {\sz SGC/RP}  & {\sz GCN/RP} \\
    \midrule
    \multirow{4}{*}{U.S.}         & {\sz income}          & \nrm{0.40} & \nrm{0.63} & \nrm{0.66} {\sz (0.46)} & \nrm{0.51} {\sz (1.0)} & \nrm{0.53} {\sz (1.3)} & \bst{0.69}       & \nrm{0.55}    & \nrm{0.55}   \\
                                  & {\sz education}       & \nrm{0.31} & \bst{0.71} & \bst{0.71} {\sz (0.00)} & \nrm{0.43} {\sz (1.0)} & \nrm{0.47} {\sz (1.0)} & \bst{0.71}       & \nrm{0.46}    & \nrm{0.48}   \\
                                  & {\fz unemployment}    & \nrm{0.47} & \nrm{0.34} & \nrm{0.39} {\sz (0.59)} & \nrm{0.32} {\sz (1.3)} & \nrm{0.45} {\sz (2.5)} & \bst{0.54}       & \nrm{0.52}    & \nrm{0.53}   \\
                                  & {\sz election}        & \nrm{0.52} & \nrm{0.42} & \nrm{0.49} {\sz (0.68)} & \nrm{0.43} {\sz (1.1)} & \nrm{0.52} {\sz (2.1)} & \bst{0.64}       & \nrm{0.61}    & \nrm{0.61}   \\
    \midrule
    \multirow{4}{*}{CDC}          & {\sz airT}            & \nrm{0.95} & \nrm{0.85} & \nrm{0.86} {\sz (0.78)} & \nrm{0.86} {\sz (2.6)} & \nrm{0.95} {\sz (3.0)} & \nrm{0.96}       & \bst{0.97}    & \bst{0.97}   \\
                                  & {\sz landT}           & \nrm{0.89} & \nrm{0.81} & \nrm{0.81} {\sz (0.09)} & \nrm{0.79} {\sz (1.0)} & \nrm{0.91} {\sz (2.4)} & \nrm{0.90}       & \nrm{0.93}    & \bst{0.93}   \\
                                  & {\sz precipitation}   & \nrm{0.89} & \nrm{0.59} & \nrm{0.61} {\sz (0.93)} & \nrm{0.61} {\sz (2.3)} & \nrm{0.79} {\sz (3.0)} & \nrm{0.89}       & \bst{0.90}    & \bst{0.90}   \\
                                  & {\sz sunlight}        & \nrm{0.96} & \nrm{0.75} & \nrm{0.81} {\sz (0.97)} & \nrm{0.80} {\sz (3.0)} & \nrm{0.90} {\sz (3.0)} & \nrm{0.96}       & \bst{0.97}    & \bst{0.97}   \\
                                  & {\sz pm2.5}           & \nrm{0.96} & \nrm{0.21} & \nrm{0.27} {\sz (0.99)} & \nrm{0.23} {\sz (2.7)} & \nrm{0.78} {\sz (3.0)} & \nrm{0.96}       & \nrm{0.96}    & \bst{0.97}   \\
    \midrule
    \multirow{4}{*}{London}       & {\sz income}          & \nrm{0.46} & \bst{0.85} & \bst{0.85} {\sz (0.00)} & \nrm{0.64} {\sz (1.0)} & \nrm{0.63} {\sz (1.0)} & \bst{0.85}       & \nrm{0.65}    & \nrm{0.64}   \\
                                  & {\sz education}       & \nrm{0.65} & \nrm{0.81} & \nrm{0.83} {\sz (0.40)} & \nrm{0.74} {\sz (1.6)} & \nrm{0.79} {\sz (1.4)} & \bst{0.86}       & \nrm{0.77}    & \nrm{0.79}   \\
                                  & {\sz age}             & \nrm{0.65} & \nrm{0.73} & \nrm{0.73} {\sz (0.17)} & \nrm{0.66} {\sz (1.2)} & \nrm{0.70} {\sz (1.7)} & \bst{0.75}       & \nrm{0.72}    & \nrm{0.72}   \\
                                  & {\sz election}        & \nrm{0.67} & \nrm{0.73} & \nrm{0.81} {\sz (0.74)} & \nrm{0.74} {\sz (2.0)} & \nrm{0.76} {\sz (2.1)} & \bst{0.85}       & \nrm{0.78}    & \nrm{0.78}   \\
    \midrule
    Twitch       & {\sz days}     & \nrm{0.08} & \nrm{0.58} & \nrm{0.59} {\sz (0.67)} & \nrm{0.22} {\sz (1.4)} & \nrm{0.26} {\sz (1.7)} & \bst{0.60}       & \nrm{0.23}    & \nrm{0.26}   \\
    \bottomrule
    \end{tabular}
    }
    \label{tab:realworld_accuracy}
\end{table*}

\xhdr{Main accuracy results}
We use the same experiment setup as we used for the synthetic datasets in \Cref{subsec:transductive_synthetic_results}, except that we no longer average over different choice of labels when reporting the performance on the same attributed graph (\cref{tab:realworld_accuracy}).
Although the vertex attributes are no longer sampled from our model, LGC still consistently outperforms SGC, 
further confirming our argument that $(\mI_{n} + \omega \mN)^{-1}$ is a more effective smoothing filter.
On the other hand, while GCN performs only slightly better than SGC on synthetic data, the extra non-linearity in GCN provides more gains over SGC for our empirical datasets (e.g., predicting pm2.5 in the CDC climate dataset).
LGC still outperforms GCN in $7$ out of $14$ prediction tasks, showing the importance of optimal feature smoothing.
Finally, the most accurate methods are the hybrid ones, as they account for both feature and outcome correlation.

\begin{figure}[t]
  \centering
  \includegraphics[width=1.00\linewidth]{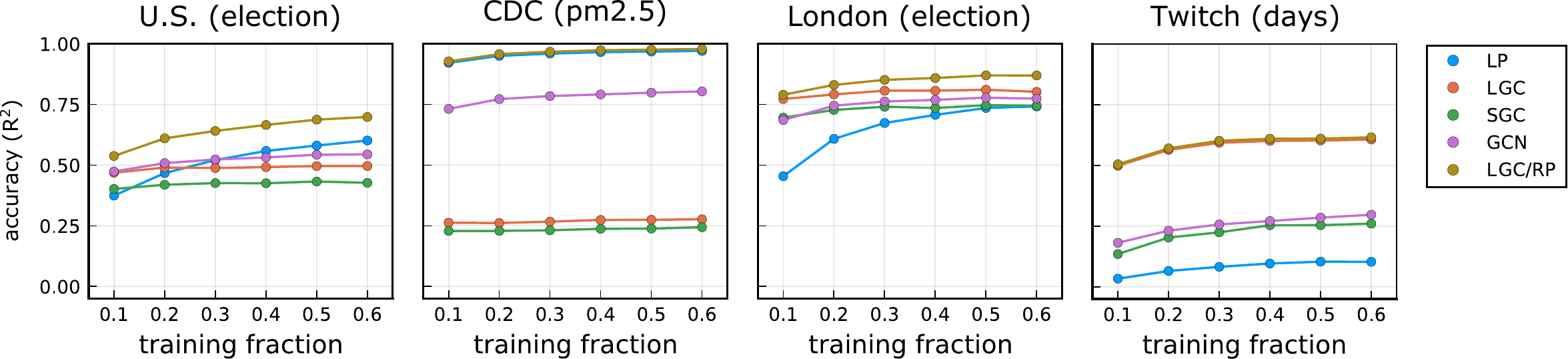}
  \caption{Transductive accuracies as a function of the fraction of nodes that are labeled.
  %
  }
\label{fig:transductive_accuracies}
\end{figure}

\xhdr{Impact of the training set size}
We pick one choice of label in each dataset and track regression accuracies as the fraction of labeled nodes increases from 10\% to 60\% (\cref{fig:transductive_accuracies}.).
We use the same experiment setup and training method, and we tune the hyperparameters on the training nodes with 5-fold cross validation.
We evaluate LP, LGC and LGC/RP, which represent three groups of algorithms based on outcome homophily, feature correlation, or both,
and also include SGC and GCN for comparison.
%
%
%
The impact of more training data varies across the algorithms.
While LP accuracy typically increases substantially with more labeled vertices, LGC and SGC accuracies are relatively constant.
Since LGC and SGC are inductive learning methods with limited capacity, they benefit little from additional examples once the training set reaches a certain size.
On the other hand, the performance of GCN moderately increase with the number of training nodes.
This is more evidence that the GCN's high model capacity allows it to partially memorize the training examples.
However, the memorization mechanism generalizes poorly to nodes in a different graph, as we demonstrate in the next section.


\section{Inductive learning experiments\label{sec:inductive_learning}}
Next, we consider the inductive learning setting, where a model is trained on one graph and tested on another.
This is useful when both graphs have similar properties, but vertex labels in one graph are difficult or expensive to obtain.
%
%

\subsection{Experiments on synthetic datasets}
\begin{figure}[t]
\centering
\includegraphics[width=0.85\linewidth]{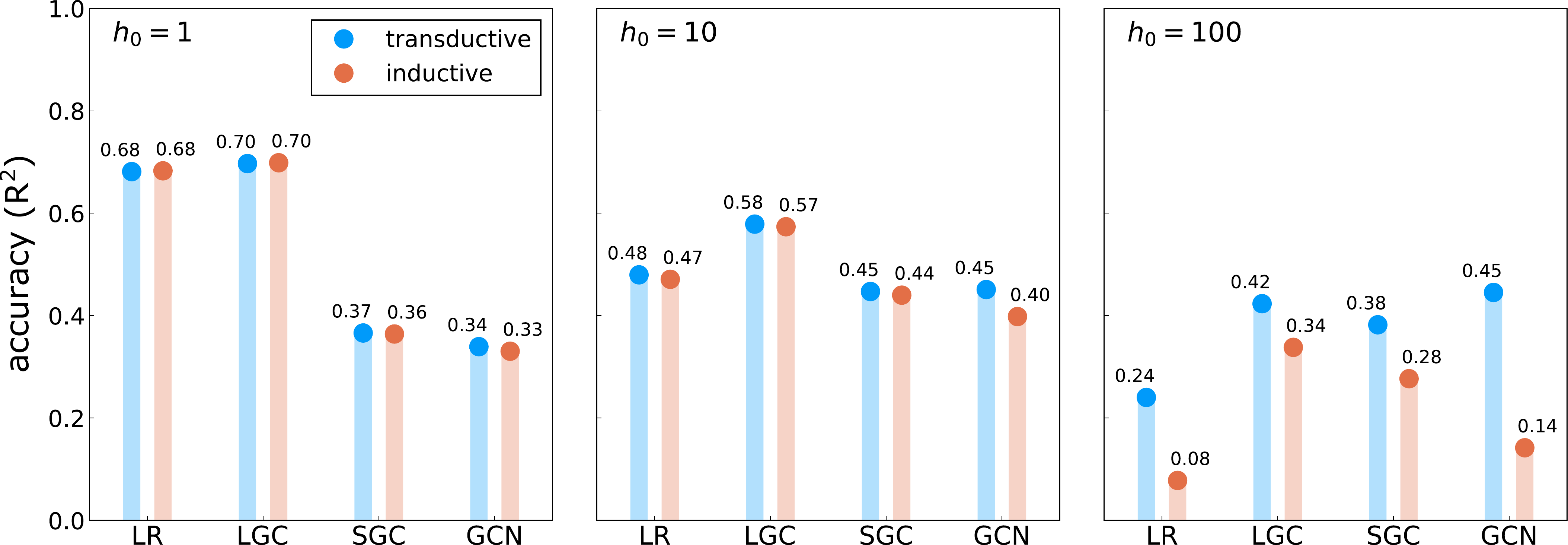}
\caption{Inductive learning accuracies on synthetic datasets sampled from our model with three $h_{0}$ values.
The transductive learning accuracies are also included for comparison.
%
%
On the attributed graph sampled with $h_{0} = 1$ or $10$, the inductive performances are close to their transductive counterparts.
When $h_{0} = 100$, outcome correlation dominates the correlation between features and outcome, and the performance gap widens;
here, the GCN outperforms LGC and SGC in transductively
but performs poorly inductively, which we argue is based on a neighborhood memorization mechanism that 
does not generalize to nodes in an unseen graph.
}
\label{fig:inductive_synthetic}
\end{figure}
The synthetic data generation uses the same Gaussian MRF as in \Cref{subsec:transductive_synthetic_results}, except that in each experiment two attributed graphs are generated, denoted $G_{1}$ and $G_{2}$.
In the transductive experiment, we used 30\% labels from $G_{1}$ for training, and the rest 70\% labels for testing.
Here, we instead use 30\% labels from $G_{2}$ for training, and test on the same 70\% labels in $G_{1}$ as the transductive case.
Since no labels in $G_{1}$ are available, we only consider LR, LGC, SGC and GCN (and not RP variants or LP).
%
%
We average the regression accuracies over repeated experiments as well as different choice of labels.
\Cref{fig:inductive_synthetic} shows the results, along with the transductive accuracies for comparison.

We expect that the inductive learning accuracies will be worse (or at best the same) compared to their transductive counterparts, 
which is exactly what we observe. 
Moreover, the gaps between inductive and transductive accuracies vary greatly as the homophily level ($h_0$) changes.
While the inductive and transductive accuracies are almost identical on graphs sampled with $h_{0}=1$, 
all four algorithms perform much worse in the inductive setting when $h_{0} = 100$.
This is because when $h_{0}$ is small, the correlation between features and outcome is the most influential, and that correlation is captured by the algorithms.
On the other hand, the outcome homophily becomes more important when $h_{0} = 100$, but none of the learning algorithms 
can take advantage of outcome correlation in $G_{1}$, as the labels are unavailable by assumption.
While the transductive accuracies of GCN is comparable to LGC and SGC, it does not generalize as well.
When $h_{0} = 100$, GCN outperforms SGC in the transductive setting but underperforms SGC by a large margin in the inductive setting.
As discussed in \Cref{subsec:transductive_realworld_results}, the GCN achieves near zero training error
and is partially memorizing the neighborhood of the training nodes.
This generalizes well to testing nodes in the same graph as they share neighbors
but not to testing nodes in a different graph.
Finally, LGC gives the highest inductive learning accuracies on all three datasets,
which is expected because the algorithm is based on optimizing for the model class from which the data was generated.

\subsection{Experiments on empirical datasets}
\begin{figure}[t]
\centering
\includegraphics[width=1.00\linewidth]{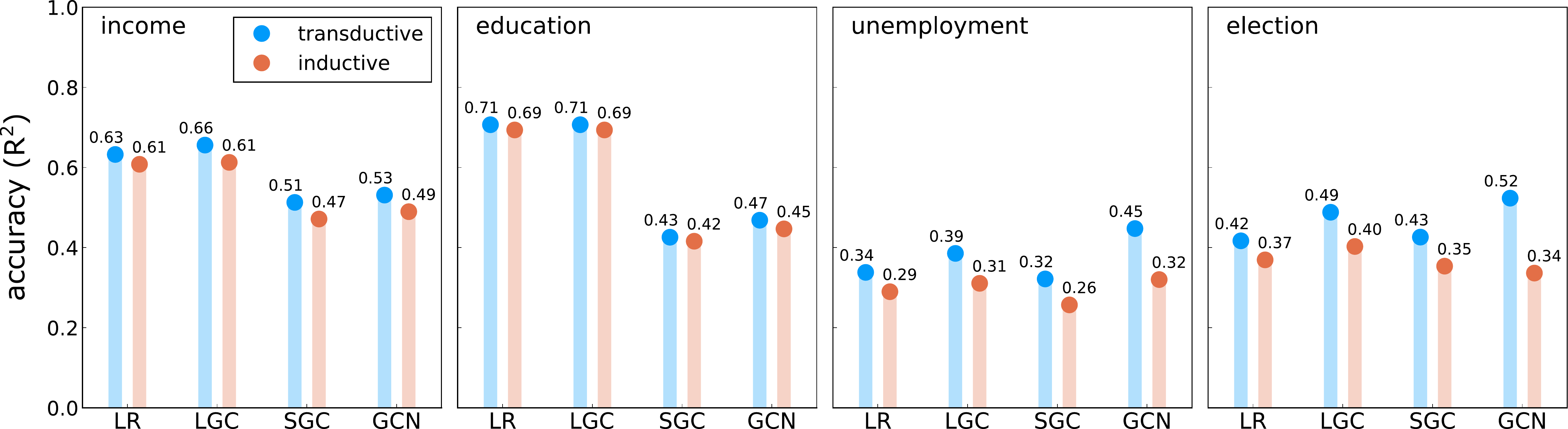}
\caption{Inductive learning accuracies for the U.S. election map with four choices of outcome.
%
%
%
%
}
\label{fig:inductive_county}
\end{figure}
Next, we compare the inductive learning capability of LR, LGC, SGC, and GCN on the U.S. election map data.
In our transductive experiments, we used data from the 2016 election.
We constructed an analogous dataset for 2012, where features and edges from the Facebook social connectedness index are identical,
but the demographic features and election outcomes are different.
The algorithms are trained on $30\%$ of labels from 2012, and tested on $70\%$ of the counties in $2016$.
We use four attributes (median household income, education level, unemployment rate and election outcome) as the label for prediction in different experiments (\cref{fig:inductive_county}).
The inductive accuracies for income and education prediction are close to the corresponding transductive accuracies, partially because those statistics are relatively stable over time, so the models trained on 2012 data provide good estimates for 2016.
When predicting attributes that changed substantially (e.g., unemployment rate and election outcome),
the GCN generalizes worse than the linear methods LGC and GCN, as shown by the larger gap between its transductive and inductive performance.
Overall, LGC achieves the highest inductive learning accuracies.

\xhdr{Interpreting LGC coefficients}
Another advantage of LGC (and linear methods in general) over GCNs is interpretability, 
as we can inspect the regression coefficients.
We average the LGC regression coefficients (i.e., $\beta$ in \cref{eq:lgc}) over different runs for the election outcome prediction,
trained on either 2012 or 2016 data, which are reported in \cref{tab:lgc_coefficients}.
Here, the election outcomes are measured by the margin of victory in each county, which can be either positive (republican won) or negative (democrat won).
There are social features (sh050m, sh100m, sh500m, for the share of Facebook friends living within 50, 100, and 500 miles),
and several demographic features.
All features are normalized to have zero mean and unit standard deviation.
From the regression coefficients using the 2016 data, 
the features with the most positive and negative coefficients are income and education.
This suggests that republican voting counties are more likely to have a higher median income and lower education level.
When comparing the regression coefficients trained on 2016 vs.\ 2012 data, the features with the largest change of coefficients are income, education and death rate.
Since a higher death rate in general indicate an older population, this suggests that republican voters in 2016 were in general richer, older, and less educated,
compared to 2012.
Finally, the social feature sh100m is associated with a large negative coefficient and sh500 a positive coefficient, 
meaning that republication voters in general might have more long-distance friends.
This aligns with the fact that many rural and sparsely populated counties lean republican in recent U.S.\ elections.
\begin{table}[t]
    \caption{LGC regression coefficients for county margin of victory in 2012 and 2016 presidential elections.
    The margin of victory is positive (republican won) or negative (democrat won).
    %
    }
    \centering
    \resizebox{1.00\linewidth}{!}{
    \begin{tabular}{r @{\qquad} ccccccccc}
    \toprule
    year  & sh050m & sh100m & sh500m & income & {\fz migration} &  birth &  death & {\fz education}  & {\fz unemployment} \\
    \midrule
    2012  &   0.06 &  -0.42 &   0.24 &   0.22 &      0.16 &  -0.13 &   0.04 &            -0.90 &  -0.38             \\
    2016  &  -0.02 &  -0.38 &   0.22 &   0.70 &      0.21 &  -0.13 &   0.51 &            -1.53 &  -0.39             \\
    \midrule
    \end{tabular}
    }
    \label{tab:lgc_coefficients}
\end{table}
\subsection{Extension to classification\label{subsec:classification}}
\begin{table}[t]
    \caption{Transductive and inductive classification accuracy on the Elliptic Bitcoin dataset as measured by the F1 score on the illicit class.
    Residual propagation is helpful in the transductive setting, and the GCN performs well.
    }
    \centering
    \resizebox{0.80\linewidth}{!}{
    \begin{tabular}{r @{\quad} c @{\quad} cccc @{\quad} ccc}
    \toprule
      Data-split &   LP &   LR & LGC &  SGC &        GCN & {\sz LGC/RP} & {\sz SGC/RP} & {\sz GCN/RP} \\
    \midrule
    transductive & 0.70 & 0.62 &   0.61 & 0.59 &      0.79  & 0.71            & 0.73         & \bst{0.83}   \\
       inductive &  --- & 0.37 &   0.31 & 0.30 & \bst{0.46} & ---             & ---          & ---          \\
    \bottomrule
    \end{tabular}
    }
    \label{tab:classification}
\end{table}
Although all the algorithms we have derived from our generative model are meant for regression problems, they can be adapted for classification by converting continuous outcome predictions to discrete classes.
In particular, for binary classification, one can simply choose a cut-off threshold for positive verses negative classes.
As a proof-of-concept example, we consider the Elliptic Bitcoin dataset~\cite{Pareja_2020}, where 
nodes represent transactions and edges correspond to payment flows.
The outcome is a ``licit'' or ``illicit'' class label for each transaction.
In this dataset, each node is associated with one of 49 time steps, where nodes from different time steps are disconnected.
This allow us to compare the inductive and transductive learning performances with different data splitting methods.
For the inductive setting, we use the same setup as Pareja et al.~\cite{Pareja_2020}, splitting nodes by time order into training/validation/testing sets with 31/5/13 time steps; 
in this setting, label propagation and residual propagation are not useful because training nodes are disconnected with testing nodes.
For the transductive setting, we use the same proportions of training, validation, testing data,
but nodes are assigned to each uniformly at random;
this setting allows us to demonstrate the effect of label homophily.

During training, we treat the prediction problem as a regression task, where we map the illicit class into a real-valued outcome $1.0$ and licit class to $0.0$.\footnote{This classification setup has well-known deficiencies~\cite{shalizi_book}, but our point here is that we can still use our methods for classification.}
For prediction, each node is first assigned a real-valued outcome estimation, then categorized into either one of the two classes depending on whether the outcome exceeds the cut-off threshold.
We treat the cut-off threshold as an additional hyperparameter, and tune it together with the others on the validation set by maximizing the F1 score for the illicit class.
Finally, we measure the classification accuracy by the F1 score for the illicit class on the testing data.
We summarize the results in \cref{tab:classification}.
First, all four algorithms (LR, LGC, SGC, and GCN) perform worse in the inductive setting than the transductive setting, as the correlation between features and labels changes over time.
Second, GCN outperforms LGC and SGC by a large margin in both settings, so nonlinear relations between features and labels plays an important role.
At last, while residual propagation does not help prediction in the inductive setting since the training and testing nodes are disconnected, it improves the classification accuracy of all the graph learning algorithms (LGC, SGC, and GCN) considerably in the transductive setting.
This experiment shows that our residual propagation post-processing can potentially help transductive node classification tasks as well.
Developing generative models similar to Gaussian MRF that directly models categorical attributes is an interesting avenue of future research.

\section{Fitting attributed graph datasets to the generative model\label{sec:model_fitting}}
So far we have focused on the algorithms derived from our Gaussian MRF model, where we used outcome homophily and feature correlation to qualitatively explain their performance differences on various datasets.
Next, we directly fit an attributed graph to our generative model.
The learned Gaussian MRF can be used to generate graphs that closely resemble the original input, or quantitatively estimate the prediction accuracies of LP, LGC, and LGC/RP.

\subsection{Maximum likelihood estimation \label{subsec:gmrf_mle}}
\begin{figure}[t]
\centering
\includegraphics[width=0.8\linewidth]{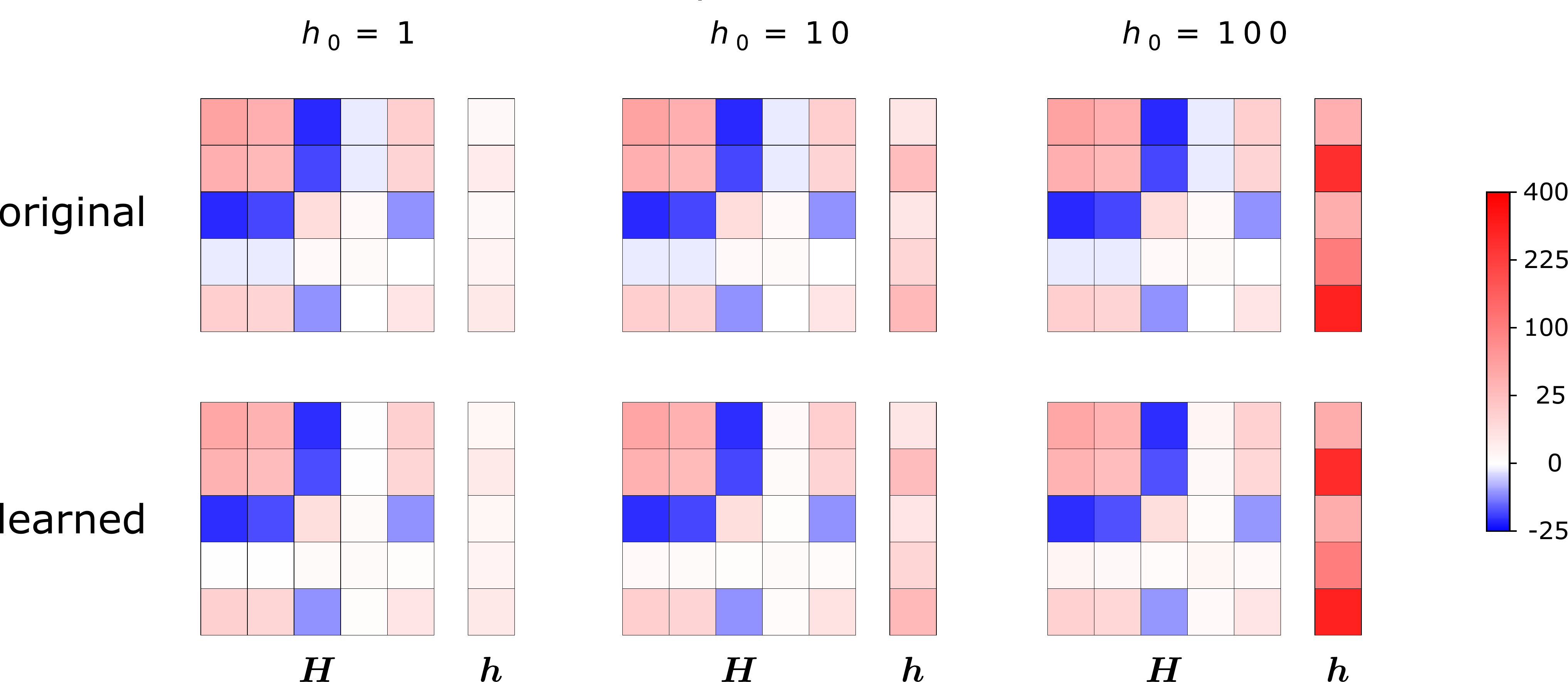}
\caption{
Visualization for the Gaussian MRF parameters $\mH \in \mathbb{R}^{5 \times 5}$ and $\vh \in \mathbb{R}^{5}$.
The top row plots the parameters used for sampling, and the bottom row plots the fitted parameters from the sampled attributed graphs.
The colormap is quadratically scaled for better visualization.
}
\label{fig:gmrf_parameters}
\end{figure}
We assume that all of the vertex attribute values are given, and our goal is to find the Gaussian MRF model that best explains how those attributes are sampled.
We use maximum likelihood estimation to fit the Gaussian MRF parameters.
The objective function is the negative log-likelihood:
\begin{align}
    \Omega(\mA, \mH, \vh) = -\log \prob(\mA | \mH, \vh) &= \left[\vectorize(\mA)^{\intercal} \mGamma \vectorize(\mA) - \log\det(\mGamma) + n(p+1)\log(2\pi)\right] / 2 \\
    &\propto \;\vectorize(\mA)^{\intercal} \mGamma \vectorize(\mA) - \log\det(\mGamma).
    \label{eq:log_likelihood}
\end{align}
The derivatives with respect to the model parameters are then
\begin{align}
\textstyle    \frac{\partial \Omega}{\partial H_{ij}} &= \textstyle   \vectorize(\mA)^{\intercal} \frac{\partial \mGamma}{\partial H_{ij}} \vectorize(\mA) - \trace\left(\mGamma^{-1} \frac{\partial \mGamma}{\partial H_{ij}}\right), &&\textstyle   \enskip \frac{\partial \mGamma}{\partial H_{ij}} = \mJ^{(ij)} \otimes \mI_{n}, \\
\textstyle       \frac{\partial \Omega}{\partial h_{i}}  &= \textstyle   \vectorize(\mA)^{\intercal} \frac{\partial \mGamma}{\partial h_{i}} \vectorize(\mA) - \trace\left(\mGamma^{-1} \frac{\partial \mGamma}{\partial h_{i}}\right), &&\textstyle   \enskip \frac{\partial \mGamma}{\partial h_{i}} = \mJ^{(ii)} \otimes \mN,
    \label{eq:log_likelihood_derivatives}
\end{align}
where $\mJ^{(ij)} \in \mathbb{R}^{(p+1) \times (p+1)}$ and $\mJ_{kl}^{(ij)} = 1$ if $kl=ij$ and $0$ otherwise.
We can then use gradient-based methods to minimize $\Omega$ with respect to $\mH$ and $\vh$.
%
However, standard matrix factorization-based algorithms for computing the log-determinant and its derivatives in \cref{eq:log_likelihood,eq:log_likelihood_derivatives} scales cubically with the number of vertices.
To exploit sparsity in $\mGamma$, we follow the algorithm outlined in our prior work~\cite{jia2020residual}, which uses conjugate gradient, stochastic trace estimation, and Lanczos quadrature to efficiently estimate the matrix log-determinant and trace of inverse~\cite{Avron_2011,Dong_2019,Fitzsimons_2018,Ubaru_2017}.
In practice, the estimation algorithm scales linearly with the number of edges in the graph and gives accurate estimation of the model parameters. 
Moreover, this procedure accurately recovers Gaussian MRF parameters, as we will show next.

\xhdr{Validation of parameter fitting}
Here, we examine the accuracy of the fitted Gaussian MRF parameters on synthetic data by comparing the estimates to the parameters used for sampling.
In particular, we use the synthetic attributed graphs from \Cref{subsec:transductive_synthetic_results} and use stochastic gradient descent\footnote{The stochastic part of the gradient descent procedure refers to the stochastic trace estimation.} to optimize $\mH$ and $\vh$,
starting from 32 random initializations, and we select the parameters that gives the highest likelihood.
We use an ADAMW optimizer for 3000 steps with a learning rate $10^{-3}$ and weight decay rate $2.5 \times 10^{-4}$.
\Cref{fig:gmrf_parameters} shows the original and the learned Gaussian MRF parameters, 
and the fitted Gaussian MRF parameters closely match the ones used to generate the data.

\subsection{Sampling data that closely resembles an input attributed graph}
\begin{figure}[t]
\centering
\includegraphics[width=1.0\linewidth]{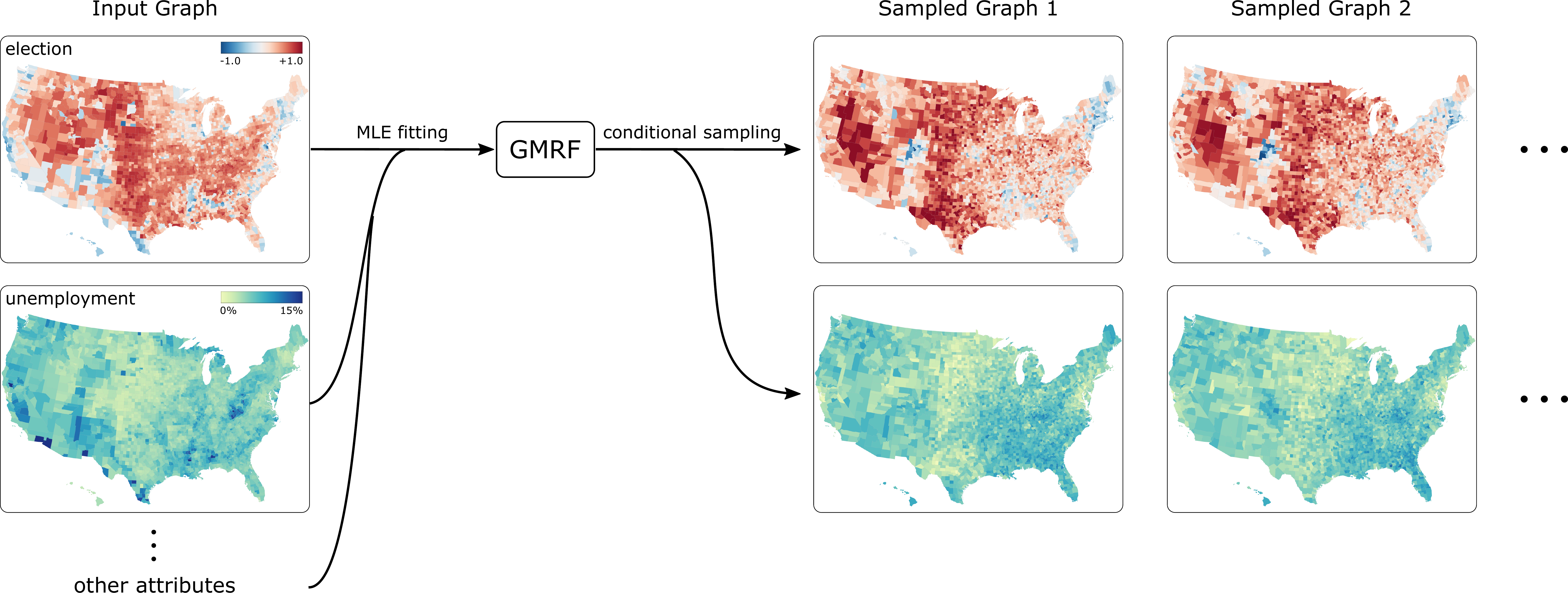}
\caption{
Election outcomes and unemployment rate that are jointly sampled from estimated Gaussian MRF, conditioned on the rest of the attributes,
where the parameters are estimated with maximum likelihood using all attributes from the U.S.\ election map dataset.
%
}
\label{fig:gmrf_sampling}
\end{figure}
Next, we fit a Gaussian MRF with the 2016 election dataset and sample attributed graphs.
This dataset consists of 10 types of vertex attributes, and we fit all attributes to the Gaussian MRF.
The learned model can be used to either jointly sample all attributes or sample some attributes while conditioning on the others.
For example, one might be interested in the influence of unemployment rate on election outcome and want to generate synthetic graphs that resemble the empirical dataset.
We sampled the joint distribution of the election outcomes and unemployment rates conditioned on all other attributes (\cref{fig:gmrf_sampling}).
In the input graph, the election outcome and unemployment rate are smooth across neighbors, and counties with higher unemployment rate tend to vote democrat.
The sampled graphs exhibit the same characteristics.

\subsection{Estimating prediction accuracy from model parameters\label{subsec:estimate_acc}}
\begin{figure}[t]
\centering
\includegraphics[width=0.93\linewidth]{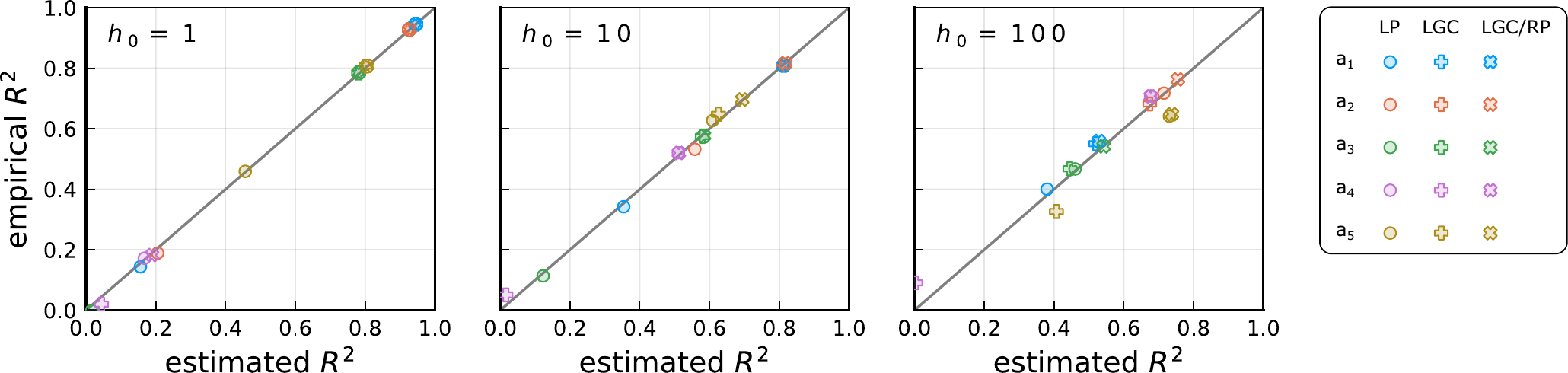}
\caption{The estimated and empirical $R^{2}$ for LP, LGC, and LGC/RP on three synthetic attributed graphs sampled from our model.
For each dataset, we run one experiment for every combination of algorithm and choice of label.
%
%
The color of each marker indicates the choice of label, and the shape indicates the algorithm.
}
\label{fig:accuracy_estimation_synthetic}
\end{figure}
\begin{figure}[t]
\centering
\includegraphics[width=1.00\linewidth]{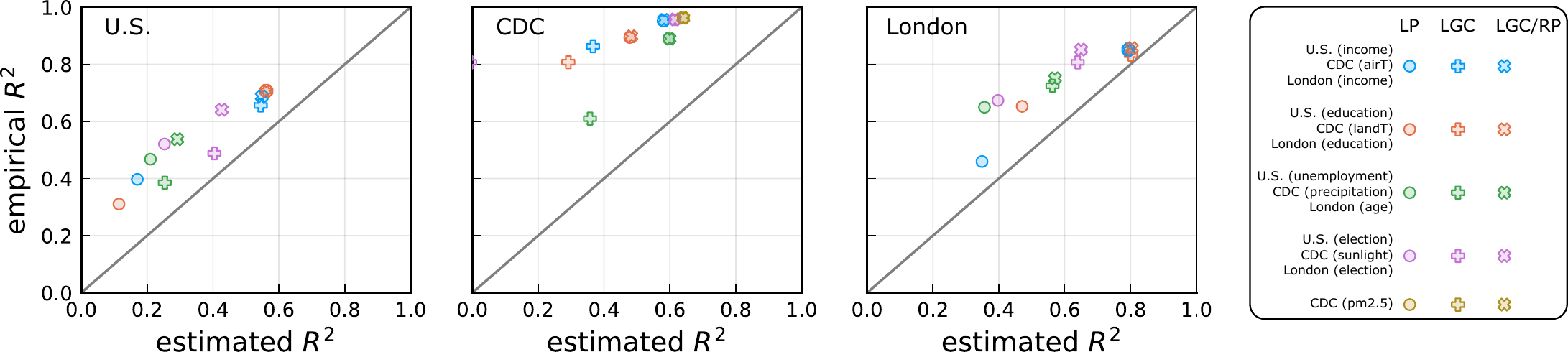}
\caption{Estimated and empirical $R^{2}$ of LP, LGC, and LGC/RP on the U.S.\ election map, CDC climate, and London election datasets.
%
%
}
\label{fig:accuracy_estimation_realworld}
\end{figure}
We saw in \Cref{sec:transductive_learning} that LP and LGC work well on different datasets, but neither consistently outperforms the other.
We have qualitatively explained this by the dominating type of correlations during the data generation process --- LP performs better when outcome homophily is more influential, while LGC performs better when feature correlation is more important.
Here, we put this idea on a quantitative footing.
First, we fit an attributed graph to the Gaussian MRF model with maximum likelihood estimation, then we can analytically estimate the regression performances of different algorithms from the Gaussian MRF parameters.

\xhdr{Estimating $R^{2}$ accuracy from Gaussian MRF}
We have previously showed LP, LGC, and LGC/RP computes the \emph{means} for some distributions of the unknown labels $\vy_{U}$, conditioned on different sets of observed attributes.
Next, we show the \emph{variance} for those conditional distributions can be used to estimate the regression accuracies for the three algorithms.
Recall the definition of our multi-attributes Gaussian MRF model from \Cref{sec:methods},
\begin{equation}
    \vectorize(\rmA) \sim \mathcal{N}(\zeros, \mSigma), \qquad \mSigma = \mGamma^{-1}, \enskip \mGamma = \mH \otimes \mI_{n} + \diagonal(\vh) \otimes \mN
\end{equation}
where $\mSigma \in \mathbb{R}^{n(p+1) \times n(p+1)}$ and $\mGamma \in \mathbb{R}^{n(p+1) \times n(p+1)}$ are the covariance matrix and precision matrix respectively.
Let $P_{L} = \{np+u : u \in L\}$ and $P_{U} = \{np+u : u \in U\}$ denote the matrix indices for nodes with observed and unobserved labels, and let
$Q = \{1 \ldots np\}$ be the indices for features.
Then, the distribution of $\vy_{U}$ conditioned on different sets of observations are as follows:
\begin{align}
    \rvy_{U} | \rvy_{L} = \vy_{L}             &\sim \mathcal{N}(\bar{\vy}_{U}^{(\rm LP)}, \mSigma^{(\rm LP)}) \\
    \rvy_{U} | \rmX     = \mX                 &\sim \mathcal{N}(\bar{\vy}_{U}^{(\rm LGC)}, \mSigma^{(\rm LGC)}) \\
    \rvy_{U} | \rvy_{L} = \vy_{L}, \rmX = \mX &\sim \mathcal{N}(\bar{\vy}_{U}^{(\rm LGC/RP)}, \mSigma^{(\rm LGC/RP)})
\end{align}
where
\begin{align}
    \mSigma^{(\rm LP)} &= \mSigma_{P_{U}P_{U}} - \mSigma_{P_{U}P_{L}}          \mSigma_{P_{L}P_{L}}^{-1}                   \mSigma_{P_{L}P_{U}} \\
    \mSigma^{(\rm LGC)} &= \mSigma_{P_{U}P_{U}} - \mSigma_{P_{U}Q}              \mSigma_{QQ}^{-1}                           \mSigma_{QP_{U}} \\
    \mSigma^{(\rm LGC/RP)} &= \mSigma_{P_{U}P_{U}} - \mSigma_{P_{U}(P_{L} \cup Q)} \mSigma_{(P_{L} \cup Q)(P_{L} \cup Q)}^{-1} \mSigma_{(P_{L} \cup Q)P_{U}}
\end{align}
Intuitively, the variance of a conditional distribution (e.g., $\rvy_{U} | \rmX = \mX$ for LGC) accounts for the remaining uncertainty in $\vy_{U}$ 
given the observation (e.g., $\rmX = \mX$) and thus serves as a lower error bound for the algorithm (e.g., LGC) that uses that observation (e.g., $\mX$) as input.
More concretely, recall that the coefficient of determination $R^{2}$ for an algorithm $\algoid$ is defined as
\begin{equation}
    R^{2}(\algoid) = 1.0 - \frac{\textrm{unexplained variance by algorithm $\algoid$}}{\textrm{total variance of $\vy_{L}$}}.
    \label{eq:r2_alg}
\end{equation}
We have argued that the unexplained variance of $\algoid$ depends on the conditional covariance matrix $\mSigma^{(\algoid)}$.
Similarly, the total variance of $\vy_{L}$ depends on the marginal distribution of $\vy_{U}$, where no information is provided:
\begin{equation}
    \rvy_{U} \sim \mathcal{N}(\bar{\vy}_{U}^{(0)}, \mSigma^{(0)}), \quad \mSigma^{(0)} = \mSigma_{P_{U}P_{U}}.
\end{equation}
Using a first order approximation for the terms in \cref{eq:r2_alg}, 
we show in \cref{subsec:R2_estimation} that the accuracy $R^{2}$ of an algorithm $\algoid$ can be estimated by
\begin{equation}
    E[R^{2}(\algoid)] \approx 1.0 - \frac{\textrm{tr}(\mSigma^{(\algoid)})}{\textrm{tr}(\mSigma^{(0)}) - \frac{1}{|U|} \ones^{\intercal} \mSigma^{(0)} \ones}.
    \label{eq:accuracy_estimation}
\end{equation}

\xhdr{Estimated $R^{2}$ on synthetic data}
First, we examine the $R^{2}$ estimations on synthetic data.
For each sampled graph, we first use the estimated Gaussian MRF parameters to reconstruct the covariance matrix.
Then, we randomly select 30\% of the nodes as training nodes $L$ and use \cref{eq:accuracy_estimation} to estimate the performance of different algorithms.
We repeat the estimation 10 times with different random splits, and summarize the results in \cref{fig:accuracy_estimation_synthetic}.
Each choice of label is shown as an individual data point.
There is high agreement between the estimated regression performance and their empirical values for every choice of label.
This is quite remarkable, as the estimated Gaussian MRF is sufficient for explaining the performance of different algorithms.

\xhdr{Estimated $R^{2}$ for real-world data}
Next, we use \cref{eq:accuracy_estimation} to estimate the regression performance on three real-world datasets --- the 2016 U.S.\ election map, London mayoral election, and CDC climate datasets.
For each dataset, we first fit the attributed graph to a Gaussian MRF and then randomly select 30\% of the nodes as training $L$ to estimate the performance of different algorithms.
We repeat each estimation 10 times with different random splits, and summarize the results in \cref{fig:accuracy_estimation_realworld}.
There is substantial positive correlation between the estimated and empirical $R^2$.
However, the estimates are worse compared to the synthetic data, as expected.
In particular, there is a large gap between the estimated and empirical $R^2$ for the CDC climate dataset.
This is likely due to non-linear correlation between different vertex attributes, since we observed in \Cref{subsec:transductive_realworld_results}
that the GCN performs well on this dataset.
On the other hand, \cref{eq:accuracy_estimation} gives much better estimations on the two election datasets, where the correlation is mostly linear.

\section{Conclusion}

There are plenty of graph-based semi-supervised learning algorithms for predicting attributes of nodes,
but there are hardly any data models for understanding or testing such algorithms.
In this paper, we have developed an attributed graph model, where attributes at nodes are sampled from a Gaussian Markov Random Field.

The benefits of our model are multifold.
First, our model facilitates the understanding of several existing algorithms, spanning
linear regression, label propagation, graph convolutions, and residual propagation;
all of these ideas can be derived from conditional expectations, for various types of conditioning.
In this sense, our model unifies a diverse set of methods, and also provides a principled way to combine the core ideas behind label propagation algorithms with the basic graph neural network methodology.
The main algorithms derived from our model are also extremely effective on real-world data.
Second, our model provides a testbed for understanding the deficiencies of existing approaches and solutions to address them,
which we showed by analyzing the low-pass filters used in feature transformations.
Third, in the transductive setting, our model provides a statistical framework for more rigorously understanding the vague concepts of over-smoothing (and under-smoothing)
that have been empirically discussed for graph neural networks.
Fourth, our model provides a concrete way to test algorithms for inductive learning, as we can simply sample two graphs from the same distribution.
This simple idea highlighted a major deficiency of graph neural networks, namely that they memorize neighborhood information
which hinders generalization when there is large amounts of homophily.
Fifth, we can also fit our model parameters to empirical data to generate similar graphs.

Moving forward, there are several interesting ways in which our model could be extended.
For instance, we assumed that the graph topology was given. An immediate extension would be to make the edges random as well, using any number of random graph models.
A challenging direction is generalizing the model to explain other graph neural network heuristics that can work well in practice.
For instance, GraphSAGE~\cite{hamilton2017inductive} concatenates and transforms vertex and neighborhood features as opposes to simply averaging them, and
a Graph Attention Network~\cite{velivckovic2018graph} learns edge weights based on features at nodes.
Designing generative models where these types of algorithmic ideas are meaningful is an interesting avenue for future research.


\bibliographystyle{unsrt}
\bibliography{main}
\appendix

\section{Mathematical background}
\subsection{Background on marginalizing and conditioning with multivariate Gaussian distributions\label{subsec:gaussian_review}}
Consider a multivarite Gaussian distribution over $n$ random variables $\rvz \sim \mathcal{N}(\bar{\vz}, \mSigma)$, where $\bar{\vz} \in \mathbb{R}^{n}$ is the mean and $\mSigma \in \mathbb{R}^{n \times n}$ is the covariance matrix.
The probability density function is
\begin{equation}
    \prob(\vz | \bar{\vz}, \mSigma) = (2\pi)^{-n/2} \det(\mSigma)^{-1/2} e^{-\frac{1}{2} (\vz - \bar{\vz})^{\intercal} \mSigma^{-1} (\vz - \bar{\vz})}.
\end{equation}
Consider any partition the random variables into two groups $(\rvz_{P}, \rvz_{Q})$, and we write the distribution in the block form
\begin{equation}
    \begin{pmatrix}
        \rvz_{P} \\
        \rvz_{Q}
    \end{pmatrix}
    \sim
    \mathcal{N}
    \left(
    \begin{bmatrix}
    \bar{\vz}_{P} \\
    \bar{\vz}_{Q}
    \end{bmatrix}
    ,\;\;
    \begin{bmatrix}
    \mSigma_{PP} & \mSigma_{PQ} \\
    \mSigma_{QP} & \mSigma_{QQ}
    \end{bmatrix}
    \right).
    \label{eq:multivariate_gaussian_covariance}
\end{equation}
The marginal distribution of $\rvz_{P}$ simply drops the irrelevant indices $Q$ from the mean and covariance.
\begin{equation}
    \rvz_{P}
    \sim
    \mathcal{N}
    \left(
    \bar{\vz}_{P}
    ,\;\;
    \mSigma_{PP}
    \right).
    \label{eq:multivariate_gaussian_marginalization_covariance}
\end{equation}
The conditional distribution of $\rvz_{P}$ is also a multivariate Gaussian distribution:
\begin{equation}
    \rvz_{P} | \rvz_{Q} = \vz_{Q}
    \sim
    \mathcal{N}
    \left(
    \bar{\vz}_{P} + \mSigma_{PQ} \mSigma_{QQ}^{-1} (\vz_{Q} - \bar{\vz}_{Q})
    ,\;\;
    \mSigma_{PP} - \mSigma_{PQ} \mSigma_{QQ}^{-1} \mSigma_{QP}
    \right).
    \label{eq:multivariate_gaussian_conditioning_covariance}
\end{equation}

Oftentimes, it is more economical to work with the precision (inverse covariance) matrix, defined as $\mGamma = \mSigma^{-1}$,
since in many cases the precision matrix is sparse while the covariance matrix is dense.
We can rewrite \cref{eq:multivariate_gaussian_covariance} with the precision matrix
\begin{equation}
    \begin{pmatrix}
        \rvz_{P} \\
        \rvz_{Q}
    \end{pmatrix}
    \sim
    \mathcal{N}
    \left(
    \begin{bmatrix}
    \bar{\vz}_{P} \\
    \bar{\vz}_{Q}
    \end{bmatrix}
    ,\;\;
    \begin{bmatrix}
    \mGamma_{PP} & \mGamma_{PQ} \\
    \mGamma_{QP} & \mGamma_{QQ}
    \end{bmatrix}^{-1}
    \right).
    \label{eq:multivariate_gaussian_precision}
\end{equation}
Using block matrix inversion, one can show that $\mSigma_{PP} = (\mGamma_{PP} - \mGamma_{PQ} \mGamma_{QQ}^{-1} \mGamma_{QP})^{-1}$, $(\mSigma_{PP} - \mSigma_{PQ} \mSigma_{QQ}^{-1} \mSigma_{QP}) = \mGamma_{PP}^{-1}$, and $\mSigma_{PQ} \mSigma_{QQ}^{-1} = - \mGamma_{PP}^{-1} \mGamma_{PQ}$.
Therefore the marginal and conditional distribution can be rewritten as
\begin{align}
    \rvz_{P}
    &\sim
    \mathcal{N}
    \left(
    \bar{\vz}_{P}
    ,\;\;
    (\mGamma_{PP} - \mGamma_{PQ} \mGamma_{QQ}^{-1} \mGamma_{QP})^{-1}
    \right),
    \label{eq:multivariate_gaussian_marginalization_precision} \\
    \rvz_{P} | \rvz_{Q} = \vz_{Q}
    &\sim
    \mathcal{N}
    \left(
    \bar{\vz}_{P} - \mGamma_{PP}^{-1} \mGamma_{PQ} (\vz_{Q} - \bar{\vz}_{Q})
    ,\;\;
    \mGamma_{PP}^{-1}
    \right).
    \label{eq:multivariate_gaussian_conditioning_precision}    
\end{align}

\section{Proofs and derivations}
\subsection{Constrained label propagation\label{subsec:lp_constrained}}
Consider running the label propagation method of Zhou et al.~\cite{Zhou_2004} starting with $\vy_{U}^{(0)} = \zeros$ and $\vy_{L}^{(0)} = \vy_{L}$, but only updating the unlabeled vertices:
\begin{equation}
    \forall u \in U,\quad y_{u}^{(t+1)} \leftarrow (1-\alpha) \cdot y_{u}^{(0)} + \alpha \cdot d_{u}^{-1/2} \sum_{v \in N_1(u)} d_{v}^{-1/2} y_{v}^{(t)}
    \label{eq:lp_constrained_simple}
\end{equation}
Let $\mS = \mD^{-1/2} \mW \mD^{-1/2}$ denote the normalized adjacency matrix, we can rewrite \cref{eq:lp_constrained_simple} as
\begin{align}
    \vy_{U}^{(t+1)} &\leftarrow (1-\alpha) \vy_{U}^{(0)} + \alpha  \mS_{U,U\cup L} \vy^{(t)} \\
                    &\leftarrow (1-\alpha) \vy_{U}^{(0)} + \alpha (\mS_{UU} \vy_{U}^{(t)} + \mS_{UL} \vy_{L}^{(t)}) \\
                    &\leftarrow \alpha \mS_{UU} \vy_{U}^{(t)} + \alpha \mS_{UL} \vy_{L},
    \label{eq:lp_constrained_compact}
\end{align}
where in the last line we use $\vy_{U}^{(0)} = \zeros$ and $\vy_{L}^{(t)} = \vy_{L}^{(0)} = \vy_{L}$.
The stationary point of \cref{eq:lp_constrained_compact} is
\begin{align}
    \vy_{U}^{(\infty)} &= (\mI - \alpha \mS)_{UU}^{-1} (\alpha \mS_{UL}) \vy_{L} \\
    &= -(\mI - \alpha \mS)_{UU}^{-1} (\mI - \alpha \mS)_{UL} \vy_{L} \\
    &= -\left[\mI + \frac{\alpha}{1-\alpha} (\mI - \mS)\right]_{UU}^{-1} \left[\mI + \frac{\alpha}{1-\alpha} (\mI - \mS)\right]_{UL} \vy_{L} \\
    &= -(\mI + \omega \mN)_{UU}^{-1} (\mI + \omega \mN)_{UL} \vy_{L},
\end{align}
which is the conditional mean in \cref{eq:lp_conditional_mean}.

\subsection{Alternative label propagation algorithm that marginalizes over features\label{subsec:alt_lp}}
In \Cref{subsec:lp}, we derived a label propagation algorithm assuming that there were no features on the vertices ($p=0$).
Here, we consider the case where the vertex features and labels are jointly sampled from a Gaussian MRF with $p+1$ attributes, where $p$ is a positive integer.
Furthermore, we assume the features are not observed, and we only have access to the ground truth labels on $L$.
The joint distribution for all the attributes is given by
\begin{equation}
    \vectorize(\rmA) \sim \mathcal{N}(\zeros, \mGamma^{-1}), \qquad \mGamma = \mH \otimes \mI_{n} + \diagonal(\vh) \otimes \mN.
\end{equation}
Let $Q = \{1 \ldots np\}$ and $P = \{np+1, \ldots, n(p+1)\}$ denote the precision matrix indices for the features and labels, respectively.
In contrast to \Cref{subsec:lgc}, where we estimate the labels conditioned on the features, now the features are not available and we need to marginalize over the corresponding random variables.
Following \cref{eq:multivariate_gaussian_marginalization_precision}, the marginal distribution of the labels is
\begin{equation}
    \rvy \sim \mathcal{N}(\zeros, \bar{\mGamma}^{-1}),
\end{equation}
where $\bar{\mGamma} = (\mGamma_{PP} - \mGamma_{PQ} \mGamma_{QQ}^{-1} \mGamma_{QP})$ is the precision matrix of the marginal distribution.
Then, we can estimate the unknown labels conditioned on the observed ones by
\begin{equation}
    \rvy_{U} | \rvy_{L} = \vy_{L} \sim \mathcal{N}(\bar{\vy}_{U}, \bar{\mGamma}_{UU}^{-1}),
\end{equation}
and the conditional mean is
\begin{equation}
    \bar{\vy}_{U} = -\bar{\mGamma}_{UU}^{-1} \bar{\mGamma}_{UL} \vy_{L}.
    \label{eq:lp_alt_conditional_mean}
\end{equation}
However, there are two problems with using \cref{eq:lp_alt_conditional_mean} in practice.
First, since the precision matrix $\mGamma$ associated with a real-world dataset is unknown, we do not know how to compute $\bar{\mGamma}$.
Second, even if $\mGamma$ is known, the marginal precision matrix $\bar{\mGamma}$ in general would not be sparse, and running label propagation with $\bar{\mGamma}$ would be computationally expensive.

\subsection{Approximating the expectation of the coefficient of determination \label{subsec:R2_estimation}}
Here, we show how to estimate the regression accuracy of LP, LGC, and LGC/RP from the joint distribution of the unknown labels and the predictors.
Assume there are three groups of variables\footnote{Here, $\rvx, \rvy, \rvz$ are general random variables, which is not limited to node features and labels.}
$\rvx, \rvy, \rvz$ sampled from a joint distribution with probability density function $\rho(\vx, \vy, \vz)$. 
Our goal is to estimate the coefficient of determination $R^{2}$ of an algorithm $\mathcal{A}$ that takes the values of $\rvx$ as input and predict $\rvy$ with the conditional expectation $E[\rvy|\rvx = \vx]$.
We define a centering vector $\bar{\rvy}$ that for every $i$, we have $\bar{y}_{i} = \frac{1}{n} \sum_{j} y_{j}$, where $n$ is the dimension of $\vy$.
Note that $\bar{\rvy}$ is a vector of random variables that depend on $\rvy$.
Then, we compute the expectation of $R^{2}(\mathcal{A})$ over the joint distribution $\rho(\vx, \vy, \vz)$.
\begin{align}
    E[R^{2}] &= 1 - \int d\vx\vy\vz\ \rho(\vx, \vy, \vz) \frac{(\vy - E[\rvy|\vx])^{\intercal} (\vy - E[\rvy|\vx])}{(\vy - \bar{\vy})^{\intercal} (\vy - \bar{\vy})} \\
    &\approx 1 - \frac{\int d\vx\vy\vz\ \rho(\vx, \vy, \vz) (\vy - E[\rvy|\vx])^{\intercal} (\vy - E[\rvy|\vx])}{\int d\vx\vy\vz\ \rho(\vx, \vy, \vz) (\vy - \bar{\vy})^{\intercal} (\vy - \bar{\vy})} \\
    &= 1 - \frac{\int d\vx\ \rho(\vx) \int d\vy\ \rho(\vy|\vx) (\vy - E[\rvy|\vx])^{\intercal} (\vy - E[\rvy|\vx])}{\int d\vy\ \rho(\vy) (\vy - \bar{\vy})^{\intercal} (\vy - \bar{\vy})}
    \label{eq:r2_expectation}
\end{align}
We use the first-order approximation of the expectation of a ratio by the expectation of the ratios.
Although this approximation may be crude in general, we still find it satisfactory for the experiments in \Cref{subsec:estimate_acc}.

In the case of Gaussian MRF, $\rho(\vx, \vy, \vz)$ is a joint Gaussian distribution,
\begin{equation}
    \rvx,\rvy,\rvz \sim \mathcal{N}(\bar{\vx}^{(0)}, \bar{\vy}^{(0)}, \bar{\vz}^{(0)}; \mSigma),
\end{equation}
and the marginal distribution $\rho(\vy)$ is the multivariate Gaussian distribution
\begin{equation}
    \rvy \sim \mathcal{N}(\bar{\vy}^{(0)}; \mSigma^{(0)}).
\end{equation}
Furthermore, the conditional distribution $\rho(\vy|\vx)$ is also a multivariate Gaussian,
\begin{equation}
    \rvy|\rvx=\vx \sim \mathcal{N}(\bar{\vy}^{(\algoid)}; \mSigma^{(\algoid)}),
\end{equation}
where $\bar{\vy}^{(\mathcal{A})} = E[\rvy|\vx]$, $\mSigma^{(\mathcal{A})}$ is independent of $\vx$, and $\algoid$ refers to the algorithm (which determines the variables $\rvx$ on which we condition).
Therefore, we can compute the numerator of \cref{eq:r2_expectation}:
\begin{align}
    numerator &= \int d\vx\ \rho(\vx) \int d\vy\ \rho(\vy|\vx) (\vy - E[\rvy|\vx])^{\intercal} (\vy - E[\rvy|\vx]) \\
              &= \int d\vx\ \rho(\vx) \sum_{i} \int d\vy\ \rho(\vy|\vx) (y_{i} - E[\mathrm{y}_{i}|\vx])^{2} \\
              &= \int d\vx\ \rho(\vx) \trace(\mSigma^{(\algoid)}) \\
              &= \trace(\mSigma^{(\algoid)}).
              \label{eq:r2_expectation_numerator}
\end{align}
And the denominator of \cref{eq:r2_expectation} is
\begin{align}
    denominator &= \int d\vy\ \rho(\vy) (\vy - \bar{\vy})^{\intercal} (\vy - \bar{\vy}) \\
                         &= \int d\vy\ \rho(\vy) [(\vy - \bar{\vy}^{(0)}) - (\bar{\vy} - \bar{\vy}^{(0)})]^{\intercal} [(\vy - \bar{\vy}^{(0)}) - (\bar{\vy} - \bar{\vy}^{(0)})] \\
                         &= \int d\vy\ \rho(\vy) (\vy - \bar{\vy}^{(0)})^{\intercal} (\bar{\vy} - \bar{\vy}^{(0)}) - 2\int d\vy\ \rho(\vy) (\vy - \bar{\vy}^{(0)})^{\intercal} (\bar{\vy} - \bar{\vy}^{(0)}) \nonumber \\
                         &\hspace{2.32in}+ \int d\vy\ \rho(\vy) (\bar{\vy} - \bar{\vy}^{(0)})^{\intercal} (\bar{\vy} - \bar{\vy}^{(0)}).
                         \label{eq:r2_expectation_denominator}
\end{align}
The first expectation in \cref{eq:r2_expectation_denominator} is
\begin{equation}
    \int d\vy\ \rho(\vy) (\vy - \bar{\vy}^{(0)})^{\intercal} (\bar{\vy} - \bar{\vy}^{(0)}) = \trace(\mSigma^{(0)}).
\end{equation}
For the second expectation in \cref{eq:r2_expectation_denominator}, note that both $\bar{\vy}$ and $\bar{\vy}^{(0)}$ are constant vectors, so
\begin{equation}
    \int d\vy\ \rho(\vy) (\vy - \bar{\vy}^{(0)})^{\intercal} (\bar{\vy} - \bar{\vy}^{(0)}) = \int d\vy\ \rho(\vy) (\bar{\vy} - \bar{\vy}^{(0)})^{\intercal} (\bar{\vy} - \bar{\vy}^{(0)}),
\end{equation}
which is the same as the third expectation:
\begin{align}
    \int d\vy\ \rho(\vy) (\bar{\vy} - \bar{\vy}^{(0)})^{\intercal} (\bar{\vy} - \bar{\vy}^{(0)}) &= \int d\vy\ \rho(\vy) \sum_{i=1}^{n} (\bar{y}_{i} - \bar{y}_{i}^{(0)})^{2} \\
    &= \int d\vy\ \rho(\vy) \frac{1}{n} \left(\sum_{i=1}^{n} y_{i} - \bar{y}_{i}^{(0)}\right) \left(\sum_{j=1}^{n} y_{j} - \bar{y}_{j}^{(0)}\right) \\
    &= \frac{1}{n} \sum_{i=1}^{n} \sum_{j=1}^{n} \int d\vy\ \rho(\vy) \left(y_{i} - \bar{y}_{i}^{(0)}\right) \left(y_{j} - \bar{y}_{j}^{(0)}\right) \\
    &= \frac{1}{n} \mathbf{1}^{\intercal} \mSigma^{(0)} \mathbf{1}.
\end{align}
To summarize,
\begin{equation}
    E[R^{2}] \approx 1 - \frac{\trace(\mSigma^{(\algoid)})}{\trace(\mSigma^{(0)}) - \frac{1}{n} \mathbf{1}^{\intercal} \mSigma^{(0)} \mathbf{1}}.
\end{equation}

\end{document}